\documentclass[letterpaper, 10 pt, journal, twoside]{ieeetran}

\usepackage{amsmath} 
\usepackage{bm}
\usepackage{amssymb}
\usepackage{siunitx}
\usepackage[pdftex]{graphicx}
\usepackage{epstopdf}
\usepackage{float}
\usepackage{cite}
\usepackage{color}
\usepackage{cancel}
\usepackage{dblfloatfix}
\usepackage{soul}
\usepackage{multirow}
\IEEEoverridecommandlockouts                              

\usepackage{epsfig} 

\newcommand{\ignore}[1]{}

\newcommand{\bbm}{\begin{bmatrix}}
\newcommand{\ebm}{\end{bmatrix}}
\newcommand{\bma}[1]{\left[\begin{array}{#1}}
\newcommand{\ema}{\end{array}\right]}

\DeclareMathAlphabet{\mbf}{OT1}{ptm}{b}{n}
\newcommand{\mbs}[1]{{\boldsymbol{#1}}}
\newcommand{\mbc}[1]{ \boldsymbol{\mathcal{#1}} } 

\newcommand{\mbsbar}[1]{{\bar{\boldsymbol{#1}}}}
\newcommand{\mbshat}[1]{{\hat{\boldsymbol{#1}}}}

\newcommand{\mbstilde}[1]{{\tilde{\boldsymbol{#1}}}}
\newcommand{\mbfdot}[1]{{\dot{\mbf{#1}}}}
\newcommand{\mbfbar}[1]{{\bar{\mbf{#1}}}}

\newcommand{\mbfcheck}[1]{{\check{\mbf{#1}}}}

\newcommand{\rframe}[1]{{\ensuremath{\mathcal{F}}_{#1}}}

\def\fdotb{{\raisebox{-0.6ex}{ \kern0.2ex\raisebox{0.8ex}{\tiny $\hspace*{-1ex}\circ$}}}}
\def\fddotb{{\raisebox{-0.6ex}{ \kern0.2ex\raisebox{0.8ex}{\tiny $\hspace*{-1ex}\circ\circ$}}}}

\newcommand{\f}{\frac}

\newcommand{\trans}{{\ensuremath{\mathsf{T}}}} 
\newcommand{\utimes}{ {\raisebox{-0.6ex}{ \kern-1.0ex\raisebox{0.6ex}{ \small $\mathsf{v}$}}} } %
\newcommand{\onehalf}{\mbox{$\textstyle{\frac{1}{2}}$}}

\newcommand{\beq}{\begin{equation}}
\newcommand{\eeq}{\end{equation}}
\newcommand{\bdis}{\begin{displaymath}}
\newcommand{\edis}{\end{displaymath}}
\newcommand{\beqarray}{\begin{eqnarray}}
\newcommand{\eeqarray}{\end{eqnarray}}
\newcommand{\beqarraynn}{\begin{eqnarray*}}
\newcommand{\eeqarraynn}{\end{eqnarray*}}

\newcommand{\cross}{\times}
\newcommand{\eye}{\mbf{1}}

\DeclareMathOperator{\Exp}{Exp}
\DeclareMathOperator{\Log}{Log}

\newcommand{\dcmbar}{\mbfbar{C}}
\newcommand{\dcmt}{\dcm^\trans}
\newcommand{\dcmbart}{\dcmbar^\trans}

\newcommand{\vel}{\mbf{v}}
\newcommand{\vbar}{\mbfbar{v}}

\newcommand{\pos}{\mbf{r}}
\newcommand{\rbar}{\mbfbar{r}}

\newcommand{\bias}{\mbf{b}}

\newcommand{\bbar}{\mbfbar{b}}

\newcommand{\bg}{\bias^g}

\newcommand{\ba}{\bias^a}
\newcommand{\babar}{\bbar^a}

\newcommand{\xibs}{\mbs{\xi}}

\newcommand{\RMIX}{\Delta\mathcal{X}}

\newcommand{\RMIvbar}{\Delta\mbfbar{v}}

\newcommand{\RMIrbar}{\Delta\mbfbar{r}}

\newcommand{\kmin}{{k-1}}

\newcommand{\grav}{\mbf{g}}
\newcommand{\Deltat}{\Delta t}
\newcommand{\zero}{\mbf{0}}

\DeclareMathAlphabet{\mbf}{OT1}{ptm}{b}{n}
\DeclareMathOperator*{\argmax}{arg\,max}
\DeclareMathOperator*{\argmin}{arg\,min}
\newcommand{\spliteq}{\nonumber \\ \spliteqindent}
\newcommand{\spliteqindent}{&\qquad\qquad\qquad}
\newcommand{\scalebrack}[1]{\left(#1\right)}

\newcommand{\diag}{{\ensuremath{\mathrm{diag}}}}
\newcommand{\dcm}{\mbf{C} }

\usepackage{amsthm}

\newtheorem{thm}{Requirement}

\title{Know What You Don't Know: Consistency in Sliding Window Filtering with Unobservable States Applied to Visual-Inertial SLAM}

\author{Daniil Lisus$^{1}$, Mitchell Cohen$^{1}$, and James Richard Forbes$^{1}$%
\thanks{Manuscript received: December, 15, 2022; Revised March, 10, 2023; Accepted April, 11, 2023.}
\thanks{This paper was recommended for publication by Editor Sven Behnke upon evaluation of the Associate Editor and Reviewers' comments.
This work was supported by McGill University, the NSERC Discovery Grant program, the Fonds de Recherche du Québec Nature et Technologies, and Denso Inc. in collaboration with the NSERC Alliance program.} 
\thanks{$^{1}$The authors are with the Department of Mechanical Engineering, McGill University, Montreal, Quebec H3A 0G4, Canada
        {\tt\footnotesize [daniil.lisus, mitchell.cohen3]@mail.mcgill.ca; james.richard.forbes@mcgill.ca}}%
\thanks{Digital Object Identifier (DOI): see top of this page.}
}

\begin{document}

\newpage
%
%
%
%
%
%
%
\def \myJournal {IEEE Robotics and Automation Letters}
\def \myDoi {10.1109/LRA.2023.3268043}
\def \myPaperSiteName {IEEE Xplore}
\def \myPaperSiteLink {https://ieeexplore.ieee.org/document/10103620}
\def \myYear {2023}

\def \myPaperCitation{D. Lisus, M. Cohen and J. R. Forbes, ``Know What You Don't Know: Consistency in Sliding Window Filtering With Unobservable States Applied to Visual-Inertial SLAM,'' in \textit{IEEE Robotics and Automation Letters}, vol. 8, no. 6, pp. 3382-3389, June 2023.}


\begin{figure*}[t]

\thispagestyle{empty}
\begin{center}
\begin{minipage}{6in}
\centering
This paper has been accepted for publication in \emph{\myJournal}. 
\vspace{1em}

This is the author's version of an article that has, or will be, published in this journal or conference. Changes were, or will be, made to this version by the publisher prior to publication.
\vspace{2em}

\begin{tabular}{rl}
DOI: & \myDoi\\
\myPaperSiteName: & \texttt{\myPaperSiteLink}
\end{tabular}

\vspace{2em}
Please cite this paper as:

\myPaperCitation

\vspace{15cm}
\copyright \myYear \hspace{4pt}IEEE. Personal use of this material is permitted. Permission from IEEE must be obtained for all other uses, in any current or future media, including reprinting/republishing this material for advertising or promotional purposes, creating new collective works, for resale or redistribution to servers or lists, or reuse of any copyrighted component of this work in other works.

\end{minipage}
\end{center}
\end{figure*}
\newpage
\clearpage
\pagenumbering{arabic} 

\maketitle

\begin{abstract}
Estimation algorithms, such as the sliding window filter, produce an estimate and uncertainty of desired states. This task becomes challenging when the problem involves unobservable states. In these situations, it is critical for the algorithm to ``know what it doesn't know'', meaning that it must maintain the unobservable states as unobservable during algorithm deployment. This letter presents general requirements for maintaining consistency in sliding window filters involving unobservable states. The value of these requirements for designing navigation solutions is experimentally shown within the context of visual-inertial SLAM making use of IMU preintegration.
\end{abstract}


\section{Introduction}\label{sec:intro}
\IEEEPARstart{R}{eal-world} applications sometimes require navigation algorithms to handle unobservable states. The unobservability of these states can arise from insufficient sensors due to weight or budget constraints, sensor failure post-deployment, or from the fundamental nature of the problem. In these cases, it is critical for the algorithm to ``know what it doesn't know'', meaning that it should not gain information about fundamentally unobservable states during runtime.

Estimation with unobservable states occurs in applications such as state estimation for legged robots \cite{Bloesch2013}, projectile trajectory estimation \cite{Roux2021}, and in simultaneous localization and mapping (SLAM) algorithms \cite{Julier2001}. A common type of SLAM algorithm, termed visual-inertial SLAM (VI-SLAM), combines one or more cameras with an inertial measurement unit (IMU). Online estimation approaches for these applications can be categorized as being either filtering- or optimization-based. Filtering-based approaches often make use of Kalman-like filters, where the state is only updated using current measurements. Conversely, optimization-based approaches, typically referred to as \textit{sliding window filters} (SWFs) or \textit{fixed-lag smoothers}, use a subset of measurements to estimate several states at once. This increases estimation quality at a cost of computation time. To reduce the computational cost of SWFs that utilize high-rate IMU measurements, \textit{IMU preintegration} \cite{Forster2017} is used by defining relative motion constraints between exteroceptive measurements.

A problem in applications with unobservable states is that traditional estimation approaches incur inconsistency. This inconsistency is attributed to the estimator falsely gaining information about unobservable states, and thus becoming overconfident in its estimates of those states. The consistency problem for estimators with unobservable states was noted in literature for VI-SLAM in \cite{Julier2001}, and attributed to a false information gain in the fundamentally unobservable directions of the estimator in \cite{Huang2010}, among others. Proposed solutions, heavily focused on VI-SLAM, include the observability constrained extended Kalman filters (OC-EKF) \cite{Bloesch2013, Liu2022, Huang2010, Hesch2014}, first-estimate Jacobian EKF (FEJ-EKF) \cite{Huang2010, Li2011}, and the robocentric problem formulation \cite{Castellanos2007, Huai2018}. With the introduction of the \textit{invariant extended Kalman filter} (IEKF) in \cite{Barrau2017}, another approach to fixing the consistency issues is to exploit a \textit{matrix Lie group} (MLG) formulation of the problem, specifically using a \textit{right-invariant} (RI) error definition. The consistency improvements associated with making use of a RI EKF are investigated in \cite{Roux2021, Song2022, Barrau2015IEKFVins, Zhang2017, Wu2017, Heo2018, Brossard2018, Brossard2019}. A lesser studied topic is the consistency of optimization-based approaches. The same information gain issue for this group of approaches is shown in \cite{Dong-Si2011}, with a FEJ-type fix shown in \cite{Dong-Si2012} and a RI-based fix shown in \cite{Huai2021}.

However, \cite{Dong-Si2011, Dong-Si2012, Huai2021}  present their solutions as being VI-SLAM-specific and analyze a specific matrix structure resulting from their problem setup. As a result, it is unclear how a different problem, such as one making use of IMU preintegration, might be handled. VI-SLAM is still chosen as the considered application in this paper, but the consistency analysis presented in Section \ref{sec:consistency_analysis} is application-agnostic.

The contribution of this letter is providing two general consistency requirements for SWF estimation involving unobservable states. The first requirement ensures that the initial algorithm formulation respects the true observability of the problem. The second requirement is generalized and clarified based on the analysis done in \cite{Huai2021}, and ensures that the true observability of the problem is maintained during runtime. Applying the requirements within a VI-SLAM algorithm results in the novel finding that using IMU preintegration leads to statistically consistent results in cases where using standard IMU integration does not.

The rest of the letter is structured as follows. Section \ref{sec:prelim} presents preliminaries. Section \ref{sec:consistency_analysis} presents the general consistency requirement for a SWF with unobservable states. Section \ref{sec:VI_SLAM} outlines the considered VI-SLAM problem, and analytically and experimentally showcases the application of the consistency requirements. Section \ref{sec:conclusion} summarizes the work done and discusses future extensions.
\vspace{-2mm}
\section{Preliminaries}\label{sec:prelim}
\vspace{-2mm}
\subsection{Lie Groups}\label{sec:MLG}
A \emph{Lie group} $G$ is a smooth manifold whose elements satisfy the group axioms \cite{microLieTheory}. The composition of two group elements $\mathcal{X}, \mathcal{Y} \in G$ is done using the $\circ$ operator, as $\mathcal{X} \circ \mathcal{Y} \in G$. For any $G$, there exists an associated Lie algebra $\mathfrak{g}$, a vector space identifiable with elements of $\mathbb{R}^m$, where $m$ is the degrees of freedom (DoF) of $G$. The \textit{exponential and logarithmic maps} are denoted $\exp: \mathfrak{g} \to G$ and $\log: G \to \mathfrak{g}$. The \textit{vee and wedge operators} are denoted $(\cdot)^\vee: \mathfrak{g} \to \mathbb{R}^m$ and $(\cdot)^\wedge: \mathbb{R}^m \to \mathfrak{g}$. For convenience, the maps $\Exp: \mathbb{R}^m \to G$ and $\Log: G \to \mathbb{R}^m$ are additionally defined. For $\mathcal{X} \in G$ and $\xibs \in \mathbb{R}^m$,
\begin{align}
    \mathcal{X} \triangleq \exp(\mbs{\xi}^\wedge) \triangleq \Exp(\mbs{\xi}), \hspace{5mm}
    \mbs{\xi} \triangleq \log(\mathcal{X})^\vee \triangleq \Log(\mathcal{X}).
\end{align}
The $\oplus : G \times \mathbb{R}^m \to G$ operator defines how elements in $G$ can be incremented by elements in $\mathbb{R}^m$. Given $\mathcal{X} \in G$ and $\mbs{\xi} \in \mathbb{R}^m$, one possible definition is $\mathcal{X} \oplus \mbs{\xi} = \mathcal{X} \circ \mathrm{Exp} \left(\mbs{\xi}\right)$. The function $\mbs{\eta}: G \times G \to \mathbb{R}^m$ defines a difference between two group elements. To take a derivative on $G$, $\mathcal{X}$ is parametrized using a small $\delta\xibs~\in~\mathbb{R}^m$, with a fixed ``nominal'' $\bar{\mathcal{X}} \in G$ as $\mathcal{X}(\delta\xibs)=\bar{\mathcal{X}} \oplus \Exp(\delta\xibs)$ \cite{Barfoot2014}. Letting ${f} : G \to N$, where $N$ is another Lie group, the Lie derivative of ${f}$ is defined as 
    \begin{align}
        \frac{D {f} \left(\mathcal{X} \right)}{D \mathcal{X}} \biggr\rvert_{\mathcal{X} = \bar{\mathcal{X}}} \triangleq \frac{\partial \mbs{\eta} \left({f} \left(\bar{\mathcal{X}} \oplus \delta\mbs{\xi} \right), \bar{\mathcal{X}} \right)}{\partial \delta\mbs{\xi}} \biggr\rvert_{\mbs{\xi} = \mbf{0}}.
    \end{align}
This letter makes use of the \textit{left group Jacobian}, denoted $\mbs{\mathcal{J}}_l$. 

This letter uses MLGs, where elements are simply $n \times n$ matrices, the $\circ$ operator is matrix multiplication, and the exponential and logarithmic maps coincide with the matrix exponential and logarithm. For elements of an MLG, the error between $\mbsbar{\mathcal{X}}$ and $\mbs{\mathcal{X}}$ is defined in a RI manner throughout, as $\mbs{\eta} \left(\mbsbar{\mathcal{X}}, \mbs{\mathcal{X}} \right) = \mathrm{Log} \left(\mbsbar{\mathcal{X}} \mbs{\mathcal{X}}^{-1} \right)$.

\vspace{-3.5mm}
\subsection{Maximum a Posteriori Estimation}
\textit{Maximum a Posteriori} (MAP) estimation aims to estimate a state ${\mathcal{X}} \in G$ that maximizes the posterior probability density function given measurements $\mbf{m}$ and a prior on the state $\check{\mathcal{X}}_0$. The estimate $\hat{\mathcal{X}}$ is given by \cite[Section~4.3]{barfoot2017state}
\begin{align}
    \hat{\mathcal{X}} &= \argmax_{{\mathcal{X}}} p({\mathcal{X}}|\check{\mathcal{X}}_0, \mbf{m}).
\end{align}
This problem can be equivalently posed as a \textit{weighted nonlinear} \textit{least-squares} (LS) problem as
\begin{align}\label{eq:batch_ls_problem}
    \hat{\mathcal{X}} &= \argmin_{{\mathcal{X}}} \onehalf \mbf{e}({\mathcal{X}})^\trans\mbf{W}\mbf{e}({\mathcal{X}}) = \argmin_{{\mathcal{X}}} J({\mathcal{X}}),
\end{align}
where $\mbf{e}({\mathcal{X}}) = \begin{bmatrix} \mbf{e}_0({\mathcal{X}})^\trans & \mbf{e}_1({\mathcal{X}})^\trans & \cdots & \mbf{e}_{n}({\mathcal{X}})^\trans \end{bmatrix}^\trans \in \mathbb{R}^{e}$ is a column matrix of stacked individual error terms with each error term relating the state to one measurement or prior, and $J$ is the objective function. Each error $\mbf{e}_\imath, \imath = 0, \ldots, n$ is assumed to be $\mbf{e}_\imath \sim \mathcal{N}(\mbf{0}, \mbs{\Sigma}_{\mbf{e}_\imath})$, where $\mbs{\Sigma}_{\mbf{e}_\imath}$ is the covariance on the error $\mbf{e}_i$. The matrix $\mbf{W}$ is a weight matrix such that $\mbf{W} = \mathrm{diag} (\mbs{\Sigma}_{\mbf{e}_0}^{-1}, \ldots, \mbs{\Sigma}_{\mbf{e}_n}^{-1})$. Any nonlinear LS solver, such as \textit{Gauss-Newton} (GN) or \textit{Levenberg Marquardt} (LM), can be used to solve \eqref{eq:batch_ls_problem}. These require the error Jacobian $\mbf{H}$ from
\begin{align}\label{eq:H_definition}
	\mbf{e}({\mathcal{X}}) &\approx \mbf{e}(\bar{\mathcal{X}}) + \frac{D \mbf{e}({\mathcal{X}})}{D {\mathcal{X}}}\biggr\rvert_{{\mathcal{X}} = \bar{\mathcal{X}}}\delta\mbs{\xi} = \mbf{e}(\bar{\mathcal{X}}) + \mbf{H}(\bar{\mathcal{X}})\delta\mbs{\mbs{\xi}},
\end{align}
where higher order terms are ignored, the linearization point $\bar{{\mathcal{X}}}$ is typically set to be the most recent state estimate, and $\delta\mbs{\xi} = \mbs{\eta}(\bar{{\mathcal{X}}}, {\mathcal{X}})$. The \textit{information matrix}, corresponding to the inverse covariance, is written as $\mbs{\Lambda} = \mbf{H}^\trans\mbf{W}\mbf{H}$. A GN update step is written $\delta \mbshat{\xi} = -\mbs{\Lambda}^{-1}\mbf{H}^\trans\mbf{W}\mbf{e} \in \mathbb{R}^m$, with the estimate updated as $\bar{\mathcal{X}} \leftarrow \bar{\mathcal{X}} \oplus \delta \mbshat{\xi}$. The final result is an estimate $\hat{\mathcal{X}}$ and covariance $\mbs{\Sigma}_\mathcal{X}$ on the $\mathbb{R}^m$ representation of $\hat{\mathcal{X}}$.

\begin{figure}[t!]
    \centering
    \includegraphics[width=0.4\textwidth]{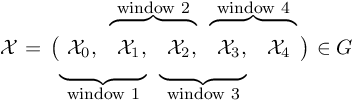}
    \vspace{-1mm}
    \caption{A visual example of the sliding window filter.}
    \label{fig:swf_example_diagram}
    \vspace{-0.5cm}
\end{figure}

Full batch estimation is often used for offline applications. For real-time applications, a recursive implementation called the SWF is used \cite{Sibley2006}. In a SWF, the problem is split into subsets called \textit{windows} as in Figure \ref{fig:swf_example_diagram}, and a batch problem is solved for each window. The process of discarding a state to move between windows is called \textit{marginalization}, where the discarded state is used to generate a prior on the new window. The details of finding this prior are found in \cite{Dong-Si2011}.

\vspace{-0.5mm}
\subsection{SLAM Formulation}
SLAM algorithms aim to estimate a map and to locate a robot within that map through time steps $t_0$ to $t_j$. In this letter, the map is represented by \emph{landmarks} $\ell^{\imath}, \imath = 0, \ldots, L$. The robot state at $t_k$ is written $\mathcal{X}_k \in G^\mathrm{rob}$. An arbitrary parametrization of the $q^\mathrm{th}$ landmark $\ell^q$ is denoted $\mathcal{Z}^q \in G^\ell$. The goal of the algorithm is to estimate the full SLAM state ${\mathcal{X}}^\mathrm{SLAM}~=~\left( \mathcal{X}_0, \hdots, \mathcal{X}_j, \mathcal{Z}^0, \hdots, \mathcal{Z}^L \right)~\in~G^{\mathrm{SLAM}}$, with covariance $\mbs{\Sigma}_\mathcal{X}$. Note, $G^{\mathrm{SLAM}}$ is a \emph{composite} group, where identity, inverse, and composition actions act on each block separately. For example, the $\circ$ operator for two SLAM states ${\mathcal{X}}^1 = \left(\mathcal{X}_0^1, \mathcal{Z}^{0_1}\right) \in G^{\mathrm{SLAM}}$ and ${\mathcal{X}}^{2} = \left(\mathcal{X}_0^2, \mathcal{Z}^{0_2}\right) \in G^{\mathrm{SLAM}}$ is defined as 
    \begin{align} 
        &{\mathcal{X}}^1 \circ {\mathcal{X}}^2 = \left(\mathcal{X}_0^1 \circ \mathcal{X}_0^2, \mathcal{Z}^{0_1} \circ \mathcal{Z}^{0_2} \right) \in G^{\mathrm{SLAM}}.
    \end{align}
The definitions of the $\oplus$ and $\mbs{\eta} \left(\cdot, \cdot \right)$ operators are similarly defined as acting component-by-component, as in \cite{microLieTheory}.

\vspace{-1mm}
\section{Consistency Analysis}\label{sec:consistency_analysis}
\subsection{Unobservability and Null Space}
An \textit{unobservable state} cannot be uniquely determined from available measurements \cite[Section~3.1.4]{barfoot2017state}. For a state with $n$ unobservable dimensions, an unobservable transformation is defined as a function  $\mathcal{T}: G \; \times \; \mathbb{R}^n \to G$, which is parameterized minimally by $\mbs{\tau} \in \mathbb{R}^n$. A transformed state is written as $\mathcal{Y}~=~\mathcal{T} \left(\mathcal{X}, \mbs{\tau} \right)$. Additionally, $\mathcal{T}$ is defined such that $\mathcal{T} \left(\mathcal{X}, \mbf{0} \right) = \mathcal{X}$. This transformation of the state does not affect the MAP objective function \eqref{eq:batch_ls_problem} \cite{Zhang2018}. To be clear, the cost functions $J \left(\mathcal{X} \right)$ and $J \left(\mathcal{Y} \right)$ are equal, while the errors $\mbf{e}(\mathcal{X})$ and $\mbf{e}(\mathcal{Y})$ are not guaranteed to be equal, contrary to what is stated in \cite{Huai2021}.

\vspace{-1mm}
\subsection{Consistency Requirements}
The form of the transformation function $\mathcal{T} \left(\mathcal{X}, \mbs{\tau} \right)$ is typically found by performing a nonlinear observability analysis, as done in \cite{Huang2010}, but in this letter, it is assumed to be known for the given problem. To analyze whether unobservable states are maintained as unobservable in a general SWF, consider the linearization of \eqref{eq:batch_ls_problem} as per \eqref{eq:H_definition}, both using the original state $\mathcal{X}$, and a transformed state $\mathcal{Y} = \mathcal{T} \left(\mathcal{X}, \mbs{\tau} \right)$. Consider $\mathcal{Y}$ linearized about some $\bar{\mathcal{X}}$ and a nominal transformation of $\mbsbar{\tau} = \zero$, meaning that $\bar{\mathcal{Y}} = \bar{\mathcal{X}}$ by the definition of $\mathcal{T}$. Comparing the general SWF objective function \eqref{eq:batch_ls_problem} for both $\mathcal{X}$ and $\mathcal{Y}$, with the weight omitted for clarity as it has no impact on the analysis, yields
\begin{align}
    \onehalf \mbf{e}(\mathcal{X})^\trans\mbf{e}(\mathcal{X}) &= \onehalf \mbf{e}(\mathcal{Y})^\trans\mbf{e}(\mathcal{Y})\label{eq:non_lin_equality}\\
    J(\bar{\mathcal{X}}) + \mbf{e}(\bar{\mathcal{X}})^\trans\mbf{H}(\bar{\mathcal{X}})\delta\mbs{\xi}^\mathcal{X} &\approx J(\bar{\mathcal{Y}}) + \mbf{e}(\bar{\mathcal{Y}})^\trans\mbf{H}(\bar{\mathcal{Y}})\delta\mbs{\xi}^\mathcal{Y} \label{eq:linearized_equality}\\
    \mbf{H}(\bar{\mathcal{X}})\delta\mbs{\xi}^\mathcal{X} &= \mbf{H}(\bar{\mathcal{Y}})\delta\mbs{\xi}^\mathcal{Y},
\end{align}
where $\delta\mbs{\xi}^\mathcal{Y} = \mbs{\eta}(\bar{\mathcal{Y}},\mathcal{Y}) = \mbs{\eta}(\bar{\mathcal{X}}, \mathcal{T} \left(\mathcal{X}, \mbs{\tau} \right) )$, $\mbf{H}(\bar{\mathcal{X}}) = \mbf{H}(\bar{\mathcal{Y}})$. The Taylor series expansion of $\delta\mbs{\xi}^\mathcal{Y}$ about $(\bar{\mathcal{X}}, \mbsbar{\tau})$ then yields
\begin{align}
    \mbf{H}(\bar{\mathcal{X}})\delta\mbs{\xi}^\mathcal{X} &= \mbf{H}(\bar{\mathcal{X}})\scalebrack{\delta\mbs{\xi}^\mathcal{X} + \underbrace{\f{\partial\mbs{\eta} \left(\bar{\mathcal{X}}, \mathcal{T} \left(\bar{\mathcal{X}}, \mbs{\tau} \right) \right) }{\partial \mbs{\tau}}\biggr\rvert_{\mbsbar{\tau} = \zero}}_{\mbf{N}}\delta\mbs{\tau}}\label{eq:N_definition}\\
    \mbf{0} &= \mbf{H} \left(\bar{\mathcal{X}} \right) \mbf{N} \left(\bar{\mathcal{X}} \right) \label{eq:general_requirement_equality},
\end{align}
where both the error Jacobian $\mbf{H} \left(\bar{\mathcal{X}} \right)$ and its null space $\mbf{N} \left(\bar{\mathcal{X}} \right)$ are evaluated at the same point $\bar{\mathcal{X}}$, and $\delta\mbs{\tau} \neq \zero$ by definition. $\mbf{N}$ has columns corresponding to each DoF of $\mbs{\tau}$, meaning that the null space of $\mbf{H}$ is spanned by vectors corresponding to each unobservable direction. Depending on the linearization approach used, the unobservable directions of the true nonlinear system may not hold after linearization. This means that not all linearization approaches will yield a valid equality going from \eqref{eq:non_lin_equality} to \eqref{eq:linearized_equality}. If \eqref{eq:general_requirement_equality} does not hold for the original linearized problem, then there is no guarantee that the problem being solved has maintained the unobservable directions through linearization. This leads to the first requirement for consistency:
\begin{thm}\label{requirement:HN_0_start}
    For a SWF algorithm with unobservable states to be consistent, \eqref{eq:general_requirement_equality} must be satisfied for any $\bar{\mathcal{X}}$. 
\end{thm}
Requirement \ref{requirement:HN_0_start} (R\ref{requirement:HN_0_start}) can be analytically checked by multiplying out each derived error Jacobian with the corresponding null space component of the states involved in the error. For linearizations that maintain R\ref{requirement:HN_0_start}, each multiplied-out equation of $\mbf{H}\mbf{N}$ will cancel to $\mbf{0}$, regardless of the evaluation points. To the best of the author's knowledge, R\ref{requirement:HN_0_start} has not previously been stated for ensuring consistency in general SWF estimation approaches with unobservable states. However, inconsistency resulting from the linearization of the nonlinear objective function is noted for EKF VI-SLAM in \cite{Li2011}, where a special, approximation-free linearization approach is taken instead. It is also hypothesized as a potential reason for why certain approximations are required for SWF VI-SLAM consistency in \cite{Huai2021}. R\ref{requirement:HN_0_start} is presented in a general way that explains the hypothesis in \cite{Huai2021}, while being applicable to other setups. Additionally, R\ref{requirement:HN_0_start} does not place restrictions on the linearization approach as is done in \cite{Li2011}, as in a general case it can be satisfied even in the presence of approximations. 

However, even with R\ref{requirement:HN_0_start} satisfied, an issue emerges during the marginalization procedure of a SWF algorithm. Consider a batch problem satisfying R\ref{requirement:HN_0_start} solved for state $\hat{\mathcal{X}}^1 = (\hat{\mathcal{X}}_m^1, \hat{\mathcal{X}}_r^1)$, where $\mathcal{X}_m^1$ is subsequently marginalized out, while $\mathcal{X}_r^1$ remains. Equation \eqref{eq:general_requirement_equality} for this problem, with all $\hat{(\cdot)}$ omitted for clarity, can be expanded as     
\begin{align}
    \zero &= \mbf{H}(\mathcal{X}^1)\mbf{N}(\mathcal{X}^1)\nonumber\\
    &= \bbm \mbf{H}_m(\mathcal{X}_m^1) & \mbf{H}_m(\mathcal{X}_r^1) \\
    \zero & \mbf{H}_r(\mathcal{X}_r^1) \ebm \bbm \mbf{N}_m(\mathcal{X}_m^1) \\ \mbf{N}_r(\mathcal{X}_r^1) \ebm,\\
    \zero &= \mbf{H}_m(\mathcal{X}_m^1)\mbf{N}_m(\mathcal{X}_m^1) + \mbf{H}_m(\mathcal{X}_r^1)\mbf{N}_r(\mathcal{X}_r^1),\\
    \zero &= \mbf{H}_r(\mathcal{X}_r^1)\mbf{N}_r(\mathcal{X}_r^1),
\end{align}
where $\mbf{H}_m$ corresponds to Jacobians involving $\mathcal{X}_m$, and $\mbf{H}_r$ corresponds to Jacobians involving only $\mathcal{X}_r$. Since $\mbf{H}_r$ does not involve $\mathcal{X}_m$, it is not present in the marginalization procedure and $\mbf{H}_r(\mathcal{X}_m) = \zero$. The $\mbf{N}_\imath$ blocks are the null space components evaluated at $\mathcal{X}_\imath, \imath = m, r$. 
Marginalization is equivalent to fixing $\mathcal{X}_m = \mathcal{X}_m^1$ and $\mbf{H}_m(\mathcal{X}) = \mbf{H}_m(\mathcal{X}^1)$, while keeping them in the problem \cite{Dong-Si2012}. When a new state subset $\mathcal{X}_n$ is added to the problem, an optimization is run again for $\mathcal{X}_r$ and $\mathcal{X}_n$, yielding a full state estimate $\hat{\mathcal{X}}^2 = (\hat{\mathcal{X}}_m^1, \hat{\mathcal{X}}_r^2, \hat{\mathcal{X}}_n^2)$. The $\mbf{H}\mbf{N}$ multiplication, with all $\hat{(\cdot)}$ omitted for clarity, is then
\begin{align}
    \bbm \mbf{H}_m(\mathcal{X}_m^1) & \mbf{H}_m(\mathcal{X}_r^1) & \zero \\
    \zero & \mbf{H}_n(\mathcal{X}_r^2) & \mbf{H}_n(\mathcal{X}_n^2) \ebm \bbm \mbf{N}_m(\mathcal{X}_m^1) \\ \mbf{N}_r(\mathcal{X}_r^2) \\
    \mbf{N}_n(\mathcal{X}_n^2)\ebm\label{eq:HN_new_prob},
\end{align}
where $\mbf{H}_n$ includes both errors used to construct $\mbf{H}_r$ and all new errors. Multiplying out \eqref{eq:HN_new_prob} produces two equations
\begin{align}
&\mbf{H}_m(\mathcal{X}_m^1)\mbf{N}_m(\mathcal{X}_m^1) + \mbf{H}_m(\mathcal{X}_r^1)\mbf{N}_r(\mathcal{X}_r^2) \spliteq = \mbf{H}_m((\mathcal{X}_m^1, \mathcal{X}_r^1))\mbf{N}_m((\mathcal{X}_m^1, \mathcal{X}_r^2))\label{eq:HN_neq_0},\\
&\mbf{H}_n(\mathcal{X}_r^2)\mbf{N}_r(\mathcal{X}_r^2) + \mbf{H}_n(\mathcal{X}_n^2)\mbf{N}_n(\mathcal{X}_n^2) \spliteq = \mbf{H}_n((\mathcal{X}_r^2, \mathcal{X}_n^2))\mbf{N}_n((\mathcal{X}_r^2, \mathcal{X}_n^2))\label{eq:HN_eq_0},
\end{align}
which are two independent $\mbf{H}\mbf{N}$ multiplications, that both need to satisfy $\zero$ to maintain \eqref{eq:general_requirement_equality}. Since $\mbf{H}$ and $\mbf{N}$ in \eqref{eq:HN_eq_0} are evaluated at the same state, \eqref{eq:HN_eq_0} is guaranteed to equal $\zero$ per R\ref{requirement:HN_0_start}. However, $\mbf{H}$ and $\mbf{N}$ in \eqref{eq:HN_neq_0} are evaluated at two different points, meaning that R\ref{requirement:HN_0_start} does not guarantee that \eqref{eq:HN_neq_0} will equal $\zero$. As a result, the null space of the problem may no longer maintain the original unobservable directions and may thus fictitiously gain information about them, leading to inconsistency. This is the ``information gain'' that has been discussed in literature, and which can be avoided by meeting the second requirement for consistency:
\begin{thm}\label{requirement:null space_constant}
    For a SWF algorithm with unobservable states satisfying R1 to be consistent, $\mbf{H} (\bar{\mathcal{X}}^1 ) \mbf{N} (\bar{\mathcal{X}}^2) = \mbf{0}$ must be satisfied for any distinct evaluation points $\bar{\mathcal{X}}^1$ and $\bar{\mathcal{X}}^2$.
\end{thm}

Requirement \ref{requirement:null space_constant} (R\ref{requirement:null space_constant}) is a generalization of the result obtained in \cite{Huai2021}. Most existing solutions to maintaining consistency with unobservable states are ensuring that R\ref{requirement:null space_constant} is satisfied. These solutions include artificially freezing Jacobian evaluation points (FEJ), projecting the Jacobian in such a way as to satisfy \eqref{eq:general_requirement_equality} for any evaluation points (OC), or ensuring that the null space is state-estimate independent (MLG-based). Jointly, R\ref{requirement:HN_0_start} and R\ref{requirement:null space_constant} are sufficient to ensure that the algorithm does not gain information in unobservable directions of the problem.
\vspace{-1mm}
\section{Consistency of VI-SLAM}\label{sec:VI_SLAM}
\subsection{Sensor Modelling}
\vspace{-0.5mm}
The considered VI-SLAM algorithm uses an IMU and a stereo camera, and involves an inertial frame $\mathcal{F}_a$, with reference point $w$, an IMU frame $\mathcal{F}_b$ with reference point $z$, and camera frames $\rframe{c_\imath}$ with reference points $c_\imath$ for the left and right camera $\imath \in [\ell, r]$. The orientation of $\mathcal{F}_b$ relative to $\mathcal{F}_a$ is denoted using the direction-cosine-matrix (DCM) $\dcm_{ab} \in SO(3)$, with $\mbf{r}_a = \dcm_{ab}\mbf{r}_b$. Frame subscripts are frequently omitted.

The IMU produces gyroscope measurements $\mbf{u}_b^g$ and accelerometer measurements $\mbf{u}_b^a$ in continuous time as
\begin{align}
	\mbf{u}_b^g(t) &= \mbs{\omega}_b(t) + \mbf{b}_b^g(t) + \mbf{w}_b^g(t),\\
	\mbf{u}_b^a(t) &= \mbf{C}_{ab}(t)^\trans(\mbf{a}_a(t)- \mbf{g}_a) + \mbf{b}_b^a(t) + \mbf{w}_b^a(t),
\end{align}
where $\mbf{g}_a$ is the gravity vector, $\mbs{\omega}_b$ is the true angular velocity, $\mbf{a}_a$ is the true acceleration, $\mbf{b}_b^\imath$ is a bias modelled with a \textit{random walk} $\mbfdot{b}_b^\imath = \mbf{w}_b^{b_\imath}$, and $\mbf{w}_b^\imath$ is \textit{white noise} with a \textit{power spectral density matrix} (PSD) $\mbc{Q}^\imath$, for $\imath \in [a, b]$.

Each camera records measurements of visible landmarks $\ell^\imath, \imath = 0, \ldots, L$, where the position of the $q^\text{th}$ landmark resolved in $\rframe{a}$ is denoted $\mbf{r}_a^{q w}$. Using a pinhole camera model, noiseless normalized image coordinates $x_n^q$ and $y_n^q$ of landmark $\ell^q$ in camera $\imath \in [\ell, r]$ are written
\begin{align}\label{eq:norm_pix_coord_proj_function}
	\mbf{g}(\mbf{r}_{c_k^\imath}^{q c_k^{\imath}}) = \begin{bmatrix} x_n^q & y_n^q\end{bmatrix}^\trans = \begin{bmatrix} x^q / z^q & y^q / z^q
	\end{bmatrix}^\trans,
\end{align}
where $\mbf{r}_{c_\imath}^{q c_\imath}$ is written
\begin{align}\label{eq:p_a_to_p_c_projection}
    \mbf{r}_{c_\imath}^{q c_\imath} = \begin{bmatrix} x^q \\ y^q \\ z^q \end{bmatrix} = \dcmt_{b c_\imath} \scalebrack{\dcmt_{ab} (\mbf{r}_a^{qw} - \pos_a^{zw}) - \mbf{t}_b^{c_\imath z}}.
\end{align}
Note that the camera-IMU \textit{extrinsic transformations} are given by $\mathcal{X}_{bc_\imath} =  (\mbf{C}_{bc_\imath}, \mbf{t}_{b}^{c_\imath z}) \in SO(3) \times \mathbb{R}^3$ and are considered known quantities in this letter. The measurements are then of the form $\mbf{y}_{q} = \mbf{g} \left(\mbf{r}_{c_\imath}^{q c_\imath} \right) + \mbf{v}_q$, where  $\mbf{v}_q \sim \mathcal{N} \left(\mbf{0}, \mbf{R}_q \right)$.
\vspace{-1mm}
\subsection{State Representation}\label{sec:state_rep}
\vspace{-1mm}
\subsubsection{Robot Representation}
It is desired to estimate the IMU orientation $\dcm_{ab} \in SO(3)$, velocity $\vel_a^{zw/a} \in \mathbb{R}^3$, position $\pos_a^{zw} \in \mathbb{R}^3$, and IMU biases $\bias_b = \bbm{\bg_b}^\trans & {\ba_b}^\trans\ebm^\trans \in \mathbb{R}^6$. These quantities are jointly referred to as the IMU state, and their parameterization belongs to the group $G^\mathrm{rob}$. In this letter, two options for $G^\mathrm{rob}$ are investigated. A standard parameterization of the IMU state is to consider each component separately, leading to $\mathcal{X}^S = \scalebrack{ \mbf{C} , \mbf{v}, \mbf{r}, \bias} \in G^S$, where the group $G^S \triangleq SO(3) \times \mathbb{R}^3 \times \mathbb{R}^3 \times \mathbb{R}^6$. The difference function for $G^S$ is given by
\begin{align}\label{eq:sep_state_def} 
    \mbs{\eta}^S(\bar{\mathcal{X}}^S, \mathcal{X}^S) &= \begin{bmatrix} \Log(\dcmbar\dcmt) \\ \vbar - \vel \\ \rbar - \pos \\ \bbar - \bias \end{bmatrix} \in \mathbb{R}^{15}.
\end{align}
Alternatively, part of the IMU state can be parameterized using the $SE_2(3)$ group \cite{Barrau2015} as
\begin{align} \label{eq:nav_state}
    \mbf{X}^{\mathrm{nav}} = \bbm \mbf{C} & \mbf{v} & \mbf{r} \\ 0 & 1 & 0 \\ 0 & 0 & 1 \ebm \in SE_2(3),
\end{align}
with the IMU state written as $\mathcal{X}^R = \scalebrack{ \mbf{X}^\mathrm{nav}, \bias } \in G^R$, where $G^R \triangleq SE_2(3) \times \mathbb{R}^6$. The difference function for $G^R$ is written as $\mbs{\eta}^R(\bar{\mathcal{X}}^R, \mathcal{X}^R) = \begin{bmatrix} \Log(\mbfbar{X}^{\mathrm{nav}} {\mbf{X}^\mathrm{nav}}^{-1})^\trans & (\bbar - \bias)^\trans \end{bmatrix}^\trans$. The $SE_2(3)$ parametrization is chosen over $SE(3) \cross \mathbb{R}^3$ as it fully exploits the group structure of the problem \cite{barrau2022geometry}.

\subsubsection{Map Representation}
The map is represented using the \textit{anchored inverse depth} parametrization as in \cite{Mourikis2007}, which represents a landmark from an \textit{anchoring state}. The parameters $\mbf{z}^q = \bbm \alpha_c^q & \beta_c^q & \lambda_c^q \ebm^\trans \in G^\ell \triangleq \mathbb{R}^3$ are estimated for each landmark $\ell^q$, and are related to their inertial position by 
\begin{align}\label{eq:world_frame_rep_from_3invdep}
	\mbf{r}_a^{qw} = \mbf{C}_i\scalebrack{\mbf{C}_{bc}\frac{1}{\lambda_c^q}\begin{bmatrix}\alpha_c^q & \beta_c^q & 1 \end{bmatrix}^\trans + \mbf{t}_{b}^{cz}} + \mbf{r}_i,
\end{align}
where $\mbf{C}_i \in SO(3)$, $\mbf{r}_i \in \mathbb{R}^3$ is the anchoring IMU orientation and position. The error function for the landmark states is written as $\mbs{\eta}^Z(\mbfbar{z}, \mbf{z}) = \scalebrack{ \bar{\alpha} - \alpha, \bar{\beta} - \beta, \bar{\lambda} - \lambda }$.
\vspace{-4mm}
\subsection{IMU Data Handling}\label{sec:imu_handling}
\subsubsection{Direct IMU Integration}
One way to use IMU data is to directly integrate the measurements to evolve the IMU state over time. The discrete-time process model given by 
\begin{align}
    \dcm_{k+1} &= \dcm_k\Exp(T_k(\mbf{u}_k^g - \bg_k - \mbf{w}_k^g)),\label{eq:one_step_proc_model_dcm}\\
    \vel_{k+1} &= \vel_k + T_k\dcm_k(\mbf{u}_k^a - \ba_k - \mbf{w}_k^a) + T_k\grav,\\
    \pos_{k+1} &= \pos_k + T_k\vel_k + \onehalf T_k^2(\dcm_k(\mbf{u}_k^a - \ba_k - \mbf{w}_k^a) + \grav),\\
    \bias_{k+1} &= \bias_k + T_k\mbf{w}_k^\bias\label{eq:one_step_proc_model_bias},
\end{align}
is used to propagate the IMU state forward. It is assumed that $\mbf{a}(t)$ and $\mbs{\omega}(t)$ are constant over the integration interval $T_k = t_{k+1} - t_k$ \cite{Forster2017}. This one-step process model can be used recursively to define a process model $f_{ij}$ linking arbitrary time steps $t_i$ to $t_j$ as
\begin{align}\label{eq:recursive_f_imu_proc_model}
   \mathcal{X}_j &= f_{j-1}\scalebrack{f_{j-2}\scalebrack{\mathcal{X}_{j-2}, \mbf{u}_{j-2}, \mbf{w}_{j-2}}, \mbf{u}_{j-1}, \mbf{w}_{j-1}}, \\
&= f_{j-1}(\mathcal{X}_{j-1}, \mbf{u}_{j-1}, \mbf{w}_{j-1})\\
  &= f_{ij}(\mathcal{X}_i, {u}_{i:j-1}, {w}_{i:j-1})
\end{align}
where ${x}_{i:j-1} = \{\mbf{x}_i, \cdots, \mbf{x}_{j-1}\}$ for $x \in [u, w]$. As the integration always occurs from an estimated state $\mathcal{X}_i$, it needs to be redone every time $\mathcal{X}_i$ is updated during optimization.

\subsubsection{IMU Preintegration}
IMU preintegration, introduced in \cite{Lupton2012} and extended to MLG theory in \cite{Forster2017}, provides a way to reduce a large number of IMU measurements into a single \textit{relative motion increment} (RMI). Given times $t_i$ and $t_j$, and defining $\Delta t_{ij} = \sum_{k=i}^{j-1} T_k$, the orientation, velocity, position, and bias RMIs are defined as
\begin{align}
	\Delta\mbf{C}_X &\triangleq \dcmt_i\dcm_j \in SO(3) ,\label{eq:ct_rmi_def_dcm_X} \\
	\Delta\mbf{v}_X &\triangleq \dcmt_i(\vel_j - \vel_i - \mbf{g}\Delta t_{ij}) \in \mathbb{R}^3, \label{eq:ct_rmi_def_vel_X}\\
	\Delta\mbf{r}_X &\triangleq \dcmt_i(\pos_j - \pos_i - \vel_i\Delta t_{ij} - \frac{1}{2}\mbf{g}\Delta t_{ij}^2) \in \mathbb{R}^3 ,\label{eq:ct_rmi_def_pos_X}\\
	\Delta\mbf{b}_X &\triangleq \bias_j - \bias_i \in \mathbb{R}^3 \label{eq:ct_rmi_def_b_X}.
\end{align}
These individual RMIs can be grouped together as $\RMIX_X$, where $\RMIX_X$ is either an element of $G^S$ or $G^R$ to match the chosen IMU state parameterization. Importantly, the RMIs can be computed incrementally from new IMU measurements as
\begin{align}
	\Delta\mbf{C}_{Y_{ij}} &\triangleq \prod_{k=i}^{j-1}\Exp\left((\mbf{u}_k^g - \bg_k - \mbf{w}_k^g) T_k\right) \in SO(3),\label{eq:dt_rmi_def_dcm_Y}\\
	\Delta\mbf{v}_{Y_{ij}} &\triangleq \sum_{k=i}^{j-1}\Delta\mbf{C}_{Y_{ik}}(\mbf{u}_k^a - \ba_k - \mbf{w}_k^a) T_k \in \mathbb{R}^3,\label{eq:dt_rmi_def_v_Y}\\
	\Delta\mbf{r}_{Y_{ij}} &\triangleq \sum_{k=i}^{j-1}\Delta\mbf{v}_{ik}T_k + \frac{1}{2}\Delta\mbf{C}_{Y_{ik}}(\mbf{u}_k^a - \ba_k - \mbf{w}_k^a) T_k^2 \in \mathbb{R}^3 \label{eq:dt_rmi_def_r_Y},\\
	\Delta\mbf{b}_{Y_{ij}} &\triangleq \mbf{0} \in \mathbb{R}^3 \label{eq:dt_rmi_def_b_Y}.
\end{align}
These RMIs are computed as a function of only the measurements and the initial bias at time $t_i$. As with $\RMIX_X$, these individual RMIs can be grouped into an element of $G^S$ or $G^R$, denoted as $\RMIX_Y$. The measured RMI $\RMIX_Y$, treated as a noisy measurement, can then be compared to $\RMIX_X$ to form an error term to be used in a batch estimation approach.

\subsubsection{Uncertainty Propagation}
Regardless of the IMU data handling approach, it is necessary to characterize the uncertainty on the resulting measurement-based quantities, either $\mathcal{X}_j$ from \eqref{eq:recursive_f_imu_proc_model} or $\RMIX_Y$ defined in \eqref{eq:dt_rmi_def_dcm_Y} to \eqref{eq:dt_rmi_def_b_Y}. As a new IMU measurement arrives at $t_\kmin \in [t_i, t_j)$, the covariance can be propagated to the next time step $t_k$, assuming $t_k - t_\kmin$ is small, using 
\begin{align} \label{eq:iterative_covariance_propagation}
	\mbs{\Sigma}_{ik} \approx \mbf{F}_{\kmin}\mbs{\Sigma}_{i\kmin}\mbf{F}_{\kmin}^\trans + \mbf{Q}_{\kmin},
\end{align}
where $\mbs{\Sigma}_{ii} = \zero$, and $\mbf{F}_{\kmin}$ and $\mbf{Q}_{\kmin}$ are the discrete-time Jacobian and process noise covariance matrices respectively.

\vspace{-0.3cm}
\subsection{Sliding Window Filter Implementation}\label{sec:swf}
\subsubsection{Algorithm Structure}
Each error $\mbf{e}_\imath$ in a SWF algorithm involves one measurement that relates a subset of IMU and/or landmark states, and which has a corresponding weight $\mbs{\Sigma}_{e_\imath}^{-1}$ and Jacobian. All these components are needed to solve the nonlinear least squares problem at each window. The considered SWF has three types of errors: a prior error that encodes the initial belief about the state; direct or preintegrated IMU errors that make use of IMU measurements; and visual errors that involve camera measurements. At each window, the oldest IMU state is marginalized out, with landmark states additionally marginalized out if the IMU state serves as their anchor. The anchoring state for a given landmark is chosen as the first IMU state at which the landmark was observed. In all presented errors, the difference function $\mbs{\eta}$ corresponds to the appropriate parameterization of the IMU state. As an example, only the direct IMU error Jacobian for the $\mathcal{X}^R$ parametrization is shown here. Expressions for all Jacobians can be found in \cite{lisusExtendedVersion}, and derivations can be found in \cite{Lisus2022}.

\subsubsection{Prior Error}
The prior error is constructed based on the marginalization procedure, which produces a mean $\check{\mathcal{X}}_{r}$ and covariance $\mbs{\Sigma}_{r}$ on a subset of the state $\mathcal{X}_{r} \in G^\mathrm{SLAM}$ that is present in both the previous and current window. The prior error is computed as $\mbf{e}_0 = \mbs{\eta}\left(\check{\mathcal{X}}_{r}, \mathcal{X}_{r} \right)$, with weight $\mbf{W}_0 = \mbs{\Sigma}_r^{-1}$, and a Jacobian with non-zero elements approximated by $\eye$ in columns corresponding to each state in $\mathcal{X}_{r}$. The approximation is made as the prior has no impact on the consistency analysis in Section \ref{sec:slam_consistency_analysis}

\subsubsection{Direct IMU Integration Error}\label{sec:imu_proc_factor}
The direct IMU integration error connects $\mathcal{X}_i$ to $\mathcal{X}_j$ using all the IMU measurements that occurred between times $t_i$ to $t_j$, where $t_i$ and $t_j$ are consecutive timesteps at which camera measurements are recorded. The error is given by $\mbf{e}_{u,j} = \mbs{\eta}(\check{\mathcal{X}}_j, \mathcal{X}_j)$, where $\check{\mathcal{X}}_j$ is computed by propagating $\mathcal{X}_i$ to time $t_j$ using \eqref{eq:recursive_f_imu_proc_model}. The weight is $\mbf{W}_{u,j} = \mbs{\Sigma}_{ij}^{-1}$ based on \eqref{eq:iterative_covariance_propagation}. Defining the Jacobian of the process model~\eqref{eq:recursive_f_imu_proc_model} with respect to the IMU state as
\begin{align}
   \f{D \check{\mathcal{X}}_j}{D \mathcal{X}_i} = \frac{D f_{ij}}{D \check{\mathcal{X}}_i},
\end{align}
the non-zero error Jacobian components are given by 
\begin{align}
    \f{D \mbf{e}_{u,j}}{D \mathcal{X}_i} &= \f{D \mbf{e}_{u,j}}{D \check{\mathcal{X}}_j}\f{D \check{\mathcal{X}}_j}{D \mathcal{X}_i} = -\mbs{\mathcal{J}}_\ell^{-1}(\mbfbar{e}_{u,j})\f{D \check{\mathcal{X}}_j}{D \mathcal{X}_i},\label{eq:e_u_jac_X_i}\\
    \f{D \mbf{e}_{u,j}}{D \mathcal{X}_j} &= \mbs{\mathcal{J}}_\ell^{-1}(-\mbfbar{e}_{u,j})\label{eq:e_u_jac_X_j},
\end{align}
with ${\mbs{\mathcal{J}}_\ell^S}^{-1} = \diag(\mbf{J}_\ell^{-1}, \eye, \eye, \zero)$ and ${\mbs{\mathcal{J}}_\ell^R}^{-1} = \diag(\mbf{J}_\ell^{-1}, \zero)$ for $\mathcal{X}^S$ and $\mathcal{X}^R$ respectively, with $\mbf{J}_\ell$ either being the $SO(3)$ or $SE_2(3)$ group Jacobian. The $\f{D \check{\mathcal{X}}_j}{D \mathcal{X}_i}$ Jacobian for $\mathcal{X}^R$, with bias-related $3\times6$ non-zero blocks marked $\star$, is
\begin{align}\label{eq:explicit_Jacobian}
    \f{D \check{\mathcal{X}}_j^R}{D \mathcal{X}_i^R} &= \bbm
                \eye & \zero & \zero & \star\\
                \Delta t_{ij}\grav_a^\wedge & \eye & \zero & \star\\
                \onehalf (\Delta t_{ij})^2\grav_a^\wedge & \Delta t_{ij}\eye & \eye & \star\\
                \zero & \zero & \zero & \eye\ebm,
\end{align}
where $\Delta t_{ij} = \sum_{k=i}^{j-1} T_k$.

\subsubsection{IMU Preintegration Error}
This error is found by comparing the measured RMI $\RMIX_Y$ defined in \eqref{eq:dt_rmi_def_dcm_Y} to \eqref{eq:dt_rmi_def_b_Y} with the predicted RMI $\RMIX_X$ defined in \eqref{eq:ct_rmi_def_dcm_X} to \eqref{eq:ct_rmi_def_b_X}, and is written $\mbf{e}_{\RMIX_{ij}} = \mbs{\eta}(\RMIX_Y, \RMIX_X)$, with weight $\mbf{W}_{u,j} = \mbs{\Sigma}_{ij}^{-1}$ per \eqref{eq:iterative_covariance_propagation}. The Jacobians are omitted for brevity.
\vspace{0.8mm}
\subsubsection{Visual Error}
In order to construct the visual error for a landmark, the landmark position $\mbf{r}_a^{qw}$ is first computed based on its parameterization using \eqref{eq:world_frame_rep_from_3invdep}. Next, \eqref{eq:norm_pix_coord_proj_function} and \eqref{eq:p_a_to_p_c_projection} are used to generate a predicted $\mbfcheck{y}_j$, which is compared to the true measurement $\mbf{y}_j$ to form the visual error $\mbf{e}_{y,j} = \mbf{y}_j - \mbfcheck{y}_j$. The weight on the error is given by $\mbf{W}_{y,j} = \mbf{R}_q^{-1}$. The Jacobians are omitted for brevity.

\vspace{-0.4cm}
\subsection{Consistency Analysis}\label{sec:slam_consistency_analysis}
\vspace{-0.1cm}
Measurements in monocular and stereo VI-SLAM algorithms are invariant to rotations of the state about the gravity vector $\phi_T$ and 3D translations in the inertial frame $\pos_T$ \cite{Huang2010}. These are the four unobservable directions of VI-SLAM, and are jointly denoted $\mbs{\tau}=\bbm \phi_T & \pos_T^\trans \ebm^\trans \in \mathbb{R}^4$. A transformation on the whole SLAM state is denoted $\mathcal{Y}^\mathrm{SLAM} = \mathcal{T}^\mathrm{SLAM}(\mathcal{X}^\mathrm{SLAM}, \mbs{\tau})$, with $\mathcal{T}^\mathrm{SLAM}$ transforming $\mathcal{X}^\mathrm{SLAM}$ component-by-component. The null space computation in \eqref{eq:general_requirement_equality} is also done in the same way. A transformation of the IMU state $\mathcal{X} = \mathcal{X}^S \in G^S$ by $\mbs{\tau}$, with the shorthand $\dcm_T \triangleq \Exp(\grav\phi_T)$, is written as
    \begin{align} \label{eq:transformation_vi_slam}
        \hspace{-0.21cm}\mathcal{Y}^X = \mathcal{T}^X (\mathcal{X}, \mbs{\tau}) = \scalebrack{\dcm_T\dcm, \dcm_T\vel, \dcm_T\pos + \pos_T, \mbf{b}_b} \in G^S.
    \end{align}
The MAP objective function is invariant for all $\mbs{\tau}$ under this transformation \cite{Zhang2018}. The transformation function in~\eqref{eq:transformation_vi_slam} can be overloaded such that with an argument $\mathcal{X} = \mathcal{X}^R \in G^R$, it modifies $\mbf{C}$, $\mbf{v}$, and $\mbf{r}$ in~\eqref{eq:nav_state} in an identical way as in~\eqref{eq:transformation_vi_slam}. The result is then rebroadcast to generate $\mathcal{Y}^X \in G^R$. Additionally, the landmark states are unaffected by $\mbs{\tau}$, as they are represented relative to an anchoring IMU state, such that $\mathcal{Y}^z = \mathcal{T}^z \left(\mbf{z}, \mbs{\tau} \right) = \mbf{z} \in \mathbb{R}^3, \; \forall \mbs{\tau} \in \mathbb{R}^4$. A prior placed on the problem shadows the unobservable directions~\cite{Dong-Si2011}. However, the algorithm should not obtain information about the unobservable directions even with a prior. Thus, the analysis is presented without a prior error.

The null spaces corresponding to $\mathcal{X}^S$ and $\mathcal{X}^R$ are
\begin{align}
    \mbf{N}_{\mathcal{X}^S} = \f{\partial\mbs{\eta}^S(\bar{\mathcal{X}}^S, \mathcal{T}^X(\bar{\mathcal{X}}^S, \mbs{\tau}))}{\partial \mbs{\tau}}\biggr\rvert_{\mbs{\tau} = \zero} = \bbm \grav & \zero\\ -\bar{\vel}^\wedge\grav & \zero \\ -\bar{\pos}^\wedge\grav & \eye \\ \zero & \zero \ebm,\label{eq:N_X_S}\\
    \mbf{N}_{\mathcal{X}^R} = \f{\partial\mbs{\eta}^R(\bar{\mathcal{X}}^R, \mathcal{T}^X(\bar{\mathcal{X}}^R, \mbs{\tau}))}{\partial \mbs{\tau}}\biggr\rvert_{\mbs{\tau} = \zero} = \bbm \grav & \zero\\ \zero & \zero \\ \zero & \eye \\ \zero & \zero \ebm\label{eq:N_X_R},
\end{align}
where $\mbf{N}_{\mathcal{X}^S}$ and $\mbf{N}_{\mathcal{X}^R}$ are both elements of $\mathbb{R}^{15 \times 4}$ and $(\cdot)^\wedge$ is the $SO(3)$ wedge operator. Of key importance is that $\mbf{N}_{\mathcal{X}^S}$ depends on $\mbfbar{v}$ and $\mbfbar{r}$, whereas $\mbf{N}_{\mathcal{X}^R}$ is state-independent. This means that R\ref{requirement:null space_constant} will be violated when $\mathcal{X}^S$ is used, as the null space component corresponding to rotation about gravity will cease to satisfy \eqref{eq:general_requirement_equality} when an evaluation point mismatch occurs. Thus, parameterizing the IMU state using the group $G^S$ will yield inconsistency regardless of any other parameter choice made.

On the other hand, the parametrization of the IMU state using the group $G^R$, exploiting the $SE_2(3)$ structure of the navigation state, satisfies R\ref{requirement:null space_constant}. Thus, if the linearization approach used satisfies R\ref{requirement:HN_0_start}, then consistency of the IMU states will be maintained. R\ref{requirement:HN_0_start} is checked analytically for both IMU state parameterizations, and for algorithms employing both direct IMU integration and IMU preintegration. For example, consider the direct IMU integration error using $\mathcal{X}^R$, with error Jacobians presented in \eqref{eq:e_u_jac_X_i} to \eqref{eq:explicit_Jacobian}. The corresponding $\mbf{H}\mbf{N}$ multiplication is
\begin{align}
    \mbf{H}\mbf{N} &= \f{D \mbf{e}_{u,j}^R}{D \mathcal{X}_i}\mbf{N}_{\mathcal{X}_i^R} + \f{D\mbf{e}_{u,j}^R}{D \mathcal{X}_j}\mbf{N}_{\mathcal{X}_j^R}\\
    &= \mbs{\mathcal{J}}_\ell^{-1}(\mbfbar{e}_{u,j}^R)\bbm
                \eye & \zero & \zero & \star\\
                \Delta t_{ij}\grav^\wedge & \eye & \zero & \star\\
                \onehalf (\Delta t_{ij})^2\grav^\wedge & \Delta t_{ij}\eye & \eye & \star\\
                \zero & \zero & \zero & \eye\ebm\bbm \grav & \zero\\ \zero & \zero \\ \zero & \eye \\ \zero & \zero \ebm \spliteq -\mbs{\mathcal{J}}_\ell^{-1}(-\mbfbar{e}_{u,j}^R)\bbm \grav & \zero\\ \zero & \zero \\ \zero & \eye \\ \zero & \zero \ebm,
\end{align}
which only equals $\mbf{0}$ if $\mbs{\mathcal{J}}_\ell^{-1} \approx \eye$, confirming the hypothesis in \cite{Huai2021}. Table \ref{tbl:sim_results} lists which combinations of state parameterizations, IMU data handling schemes, and approximations of group Jacobians satisfy R\ref{requirement:HN_0_start} and R\ref{requirement:null space_constant}. Notably, R\ref{requirement:HN_0_start} is met for all combinations making use of preintegration, including those where it is not met with direct integration. Indeed, the way that linearization is handled in \cite{Li2011} is very similar to preintegration. This makes IMU preintegration not only computationally cheaper, but also easier to achieve consistency with in the context of VI-SLAM. This conclusion, and the way it is reached by making use of the two general requirements, is the major contribution of this letter. Note that for other problems, finding suitable state representations to achieve consistency can be done by considering the geometry of the problem at hand \cite{barrau2022geometry}.

Finally, noting that $\mathcal{T}^z \left(\mbf{z}, \mbs{\tau} \right) = \mbf{z}$, the null space component for the landmark states is $\mbf{N}_z = \zero$, meaning that R1 and R2 are automatically satisfied for landmark components of $\mbf{H}\mbf{N}$. The parts of $\mbf{H}\mbf{N}$ corresponding to derivatives of visual errors with respect to IMU states are verified to fulfill R1 for all considered parametrizations.
\begin{figure}[t]
    \centering
    \includegraphics[width=0.45\textwidth, trim={5cm 0.1cm 5cm 0},clip]{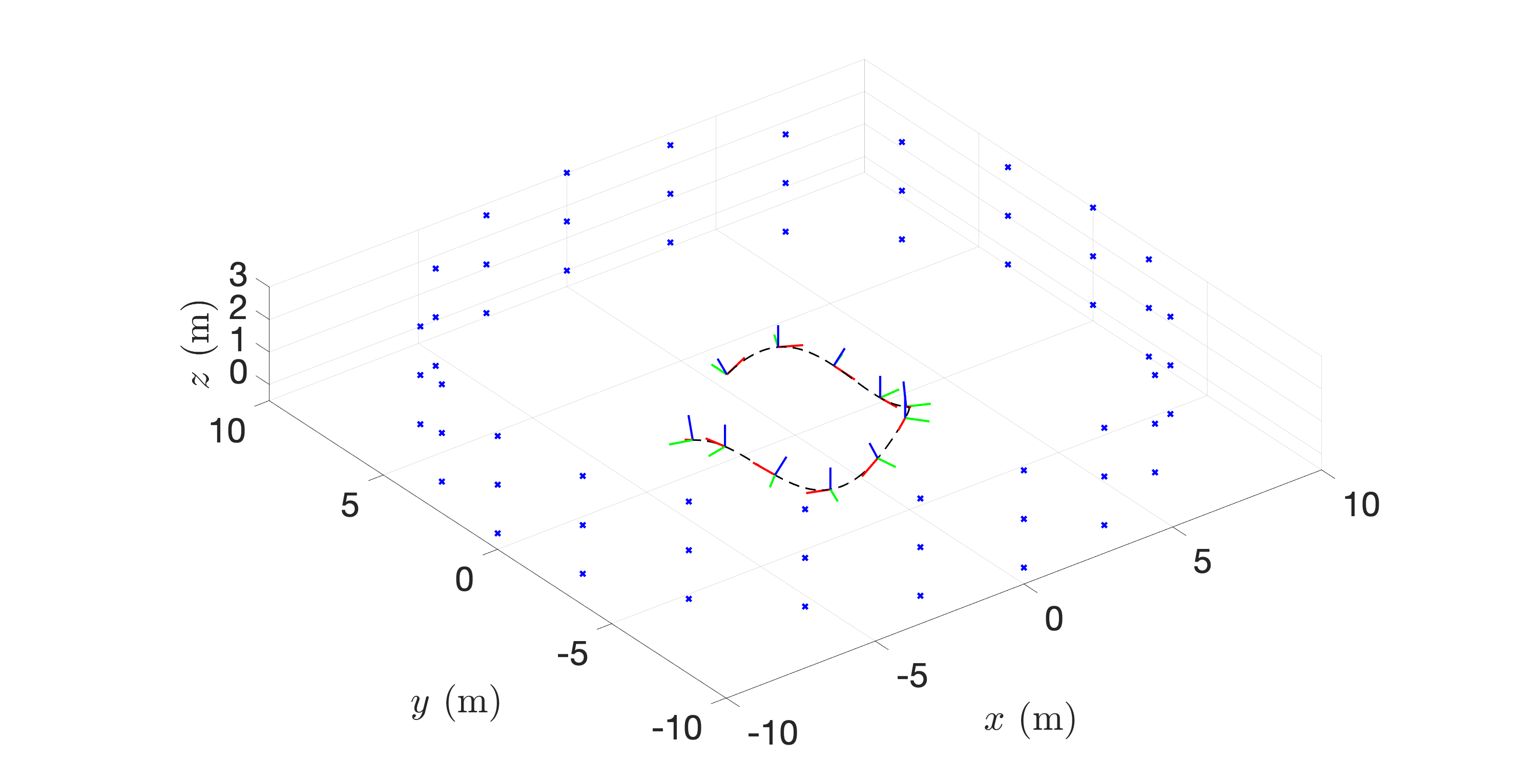}
    \vspace{-3mm}
    \caption{A snippet of the simulated trajectory with the robot body frame visualized and landmarks in blue.}
    \label{fig:traj_visualization}
\end{figure}
\subsection{Empirical Evaluation}\label{sec:results}
\begin{figure*}[t!]
    \centering
    \includegraphics[width=\textwidth, trim={4cm 0 4cm 0},clip]{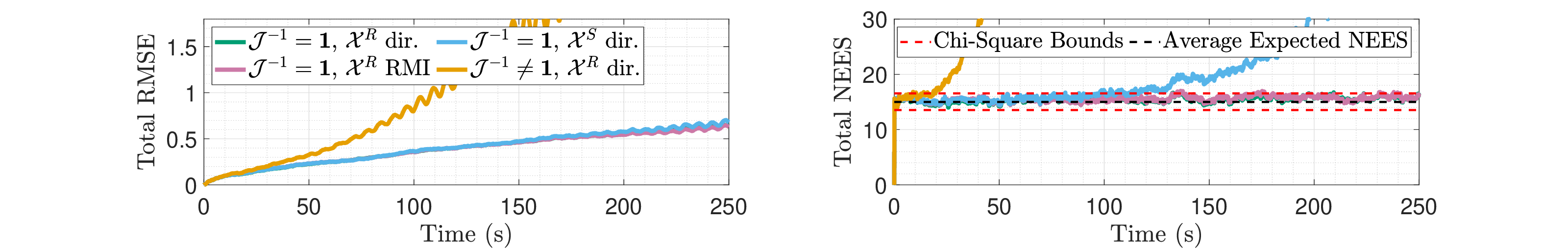}
    \vspace{-0.4cm}
    \caption{Select simulation RMSE and NEES results. Labels of ``dir.'' and ``RMI'' refer to direct IMU integration and preintegration respectively. The lines for $\mbs{\mathcal{J}}^{-1}=\eye$, $\mathcal{X}^R$ dir. and $\mbs{\mathcal{J}}^{-1}=\eye$, $\mathcal{X}^R$ RMI are overlapping almost perfectly, with the two indistinguishable in the left plot.}
    \label{fig:sim_results}
\end{figure*}
\renewcommand{\arraystretch}{1.25}
\begin{table*}[t!]
\centering
\begin{tabular}{||c|c|c|c|c|c|c|c|c|c|c|c||}
\hline
\hline
\multirow{2}{*}{IMU Handling} & \multirow{2}{*}{State} & \multirow{2}{*}{$\mbs{\mathcal{J}}_\ell^{-1} \approx \eye$} & \multirow{2}{*}{R\ref{requirement:HN_0_start}} & \multirow{2}{*}{R\ref{requirement:null space_constant}} & \multicolumn{3}{c|}{Average RMSE} & \multicolumn{3}{c|}{Average NEES} & \multirow{2}{*}{Consistent}\\
\cline{6-11}
& & & & & Total ($\sim$) & Yaw (\si{\degree}) & Position (\si{\metre}) & Total (15) & Yaw (1) & Position (3) & \\
\hline
\hline
    & \multirow{2}{*}{$\mathcal{X}^R$} & \textbf{Yes} & \textbf{Yes} & \textbf{Yes} & \textbf{0.39} & \textbf{4.13} & \textbf{0.85} & \textbf{15.47} & \textbf{1.17} & \textbf{3.21} & \textbf{Yes}\\
    \cline{3-12}
    \multirow{2}{*}{Direct} & & No & No & No & 1.42 & 60.06 & 2.61 & 1727.51 & 1088.93 & 119.97 & No\\
    \cline{2-12}
    \cline{2-12}
    & \multirow{2}{*}{$\mathcal{X}^S$} & Yes & Yes & No & \textbf{0.40} & 5.35 & \textbf{0.86} & 22.07 & 4.95 & \textbf{3.45} & No\\
    \cline{3-12}
    & & No & No & No & \textbf{0.39} & 4.71 & \textbf{0.85} & 24.68 & 8.23 & \textbf{3.43} & No\\
    \hline
    \hline
    & \multirow{2}{*}{$\mathcal{X}^R$} & \textbf{Yes} & \textbf{Yes} & \textbf{Yes} & \textbf{0.39} & \textbf{4.16} & \textbf{0.85} & \textbf{15.57} & \textbf{1.17} & \textbf{3.22} & \textbf{Yes}\\
    \cline{3-12}
    \multirow{2}{*}{Preintegration} & & \textbf{No} & \textbf{Yes} & \textbf{Yes} & \textbf{0.39} & \textbf{4.16} & \textbf{0.85} & \textbf{15.64} & \textbf{1.18} & \textbf{3.25} & \textbf{Yes}\\
    \cline{2-12}
    \cline{2-12}
    & \multirow{2}{*}{$\mathcal{X}^S$} & Yes & Yes & No & \textbf{0.40} & 5.35 & \textbf{0.86} & 22.07 & 4.94 & \textbf{3.47} & No\\
    \cline{3-12}
    & & No & Yes & No & 0.41 & 5.05 & 0.88 & 20.81 & 4.25 & 3.44 & No\\
\hline
\hline
\end{tabular}
\vspace{2mm}
\caption{Simulation RMSE and NEES results. Best RMSE (within 5\%), consistent NEES, and best performing selections for each IMU approach are bold. Fulfilment of Requirements \ref{requirement:HN_0_start} and \ref{requirement:null space_constant} (R\ref{requirement:HN_0_start} \& R\ref{requirement:null space_constant}) is indicated. Expected avg. NEES is shown in brackets.}
\label{tbl:sim_results}
\vspace{-0.6cm}
\end{table*}
\subsubsection{Simulation}
A robot moving for 250 seconds in a circular 3D sine wave while observing 60 landmarks equally spaced in a cylinder around the trajectory is simulated using \texttt{MATLAB}. A snippet of this trajectory is visualized in Figure~\ref{fig:traj_visualization}. To remove the influence of discretization errors, the ground truth states are generated using \eqref{eq:one_step_proc_model_dcm} to \eqref{eq:one_step_proc_model_bias}. Noise is modelled as Gaussian and equal in all axes, with standard deviation of 0.01 \si{\radian\per\second\squared} and \si{\meter\per\second\squared} for the IMU noises, 0.001 \si{\radian\per\second\cubed} and \si{\meter\per\second\cubed} for the random walks, and 1 pixel for the camera. The IMU and camera have frequencies of 200 \si{\Hz} and 10 \si{\Hz} respectively. Gravity is set to $\mbf{g}_a = \bbm 0 & 0 & -9.81 \ebm^\trans$ \si{\metre\per\second\squared}. Biases are initialized to $\zero$.  The stereo camera is composed of two identical cameras with intrinsics $f_u = f_v = 385.75$, $c_u = 323.12$, $c_v = 236.74$, a $640 \times 480$ pixel plane, and a baseline of $0.15$ \si{\metre}. All presented results are from 100 Monte Carlo trials with randomized noise and using the LM algorithm. The averaged \textit{root mean squared error} (RMSE) and \textit{normalized estimation error squared} (NEES) values are presented for the full IMU state, the yaw component, and the position component. The NEES bounds and expected value for a $99.7\%$ confidence bound based on the chi-square test \cite[Section~5.4.2]{bar2004estimation} for select setups are visualized in Figure~\ref{fig:sim_results}, with all results summarized in Table \ref{tbl:sim_results}.

Overall, the results confirm the consistency analysis. Only the $\mathcal{X}^R$ parametrization is able to produce overall consistent results. The $\mathcal{X}^S$ parametrization yields consistent position results, since information is only gained in the rotation about gravity component. The sufficiency of R\ref{requirement:HN_0_start} is also confirmed: results that meet both requirements achieve overall consistency. Finally, as expected, the RMSE is near identical regardless of which IMU data handling scheme is used, since the information used for state estimation is not impacted by the data handling scheme. Notably, preintegration is able to achieve consistency even when $\mbs{\mathcal{J}}_\ell^{-1}$ is not approximated by $\eye$ with the $\mathcal{X}^R$ parametrization, as predicted by the theory.

\subsubsection{EuRoC Data}\label{sec:euroc_results} 
To validate the theory on real data, five machine hall trajectories from the EuRoC micro aerial vehicle dataset are used \cite{Burri2016}. IMU and camera parameters are set based on the included parameter file for each set. The KLT-based feature tracker from VINS-Fusion \cite{Qin2019} is used to generate feature measurements from the raw images. As handling real world complexities such as imperfect knowledge of camera intrinsics and extrinsics, and data association is outside of the scope of this letter, the EuRoC data results are included as a relative performance metric. Due to the real world complexities, it is not expected that any of the results will be consistent. However, it is hypothesized that the more statistically consistent setups should achieve more consistent results, meaning that the NEES values should be closer to the expected average value. For brevity, only the results where IMU preintegration is used and $\mbs{\mathcal{J}}^{-1}\approx\eye$ are shown. Numerical results are presented in Table \ref{tbl:euroc_results}

Minimal difference is seen between using direct integration and preintegration and, with inconsistency arising from many real world factors, the group Jacobian approximation has a negligible effect. Inconsistencies stemming from real world factors appear to be much more significant than those that arises from improperly maintaining observable directions. However, the low RMSE across all results implies that state estimates are reliable despite their inconsistency. The $\mathcal{X}^R$ parametrization does produce as good as (within 5\%) or better NEES values compared to the $\mathcal{X}^S$ parametrization. The performance improvement is not as clear in the more difficult mh\_04 and mh\_05 sets, which can be explained by the presence of more real world inconsistencies due to the degraded lighting and motion conditions. The EuRoC results highlight that much more work needs to be done in real world applications to achieve full consistency. Nonetheless, the simulation results show that if all other sources of inconsistency are properly managed, consistency can be ensured by properly handling unobservable states using R\ref{requirement:HN_0_start} and R\ref{requirement:null space_constant}.

\section{Conclusion}\label{sec:conclusion}
This letter presents two requirements for consistency in a SWF algorithm involving unobservable states. These requirements are applied to a VI-SLAM problem and are used to explain the consistency properties of the algorithm using different state parametrizations and data handling. The requirements are leveraged to show that it is possible, and indeed easier, to get consistent results for a SWF VI-SLAM algorithm using IMU preintegration, a novel contribution. The VI-SLAM algorithm is also applied to real data, where it is shown that external sources of inconsistency are more impactful than the information gain from the algorithm setup. However, while it is possible to eliminate external sources of inconsistency, meeting the two presented requirements is still crucial for a SWF algorithm to maintain unobservable states as unobservable.

\begin{table*}
\centering
\begin{tabular}{||c|c|c|c|c|c|c|c||}
\hline
\hline
\multirow{2}{*}{Set} & \multirow{2}{*}{State} & \multicolumn{3}{c|}{Average RMSE} & \multicolumn{3}{c|}{Average NEES}\\
\cline{3-8}
& & Total ($\sim$) & Yaw (\si{\degree}) & Position (\si{\metre}) & Total (15) & Yaw (1) & Position (3)\\
\hline
\hline
    \multirow{2}{*}{mh\_01} & $\mathcal{X}^R$ & \textbf{0.09} & \textbf{1.80} & \textbf{0.17} & \textbf{1065.95} & \textbf{69.33} & \textbf{160.38}\\
    \cline{2-8}
    \cline{2-8}
    & $\mathcal{X}^S$ & \textbf{0.09} & \textbf{1.81} & \textbf{0.17} & \textbf{1084.36} & 77.1 & 362.78\\
\hline
\hline
    \multirow{2}{*}{mh\_02} & $\mathcal{X}^R$ & \textbf{0.05} & \textbf{0.34} & \textbf{0.11} & \textbf{548.45} & \textbf{4.55} & \textbf{153.61}\\
    \cline{2-8}
    \cline{2-8}
    & $\mathcal{X}^S$ & \textbf{0.05} & \textbf{0.34}& \textbf{0.11} & \textbf{553.26} & \textbf{4.76} & 213.31\\
\hline
\hline
    \multirow{2}{*}{mh\_03} & $\mathcal{X}^R$ & \textbf{0.06} & \textbf{0.48} & \textbf{0.14} & \textbf{829.55} & \textbf{9.72} & \textbf{177.74}\\
    \cline{2-8}
    \cline{2-8}
    & $\mathcal{X}^S$ & \textbf{0.06} & \textbf{0.47} & \textbf{0.14} & \textbf{839.80} & \textbf{9.54} & \textbf{183.89}\\
\hline
\hline
    \multirow{2}{*}{mh\_04} & $\mathcal{X}^R$ & \textbf{0.09} & 0.68 & \textbf{0.20} & \textbf{841.38} & 19.79 & \textbf{130.86}\\
    \cline{2-8}
    \cline{2-8}
    & $\mathcal{X}^S$ & \textbf{0.09} & \textbf{0.58} & \textbf{0.20} & \textbf{846.28 }& \textbf{16.58} & 257.68\\
\hline
\hline
    \multirow{2}{*}{mh\_05} & $\mathcal{X}^R$ & \textbf{0.07} & \textbf{0.49} & \textbf{0.14} & \textbf{793.50} & 12.32 & \textbf{210.06} \\
    \cline{2-8}
    \cline{2-8}
    & $\mathcal{X}^S$ & \textbf{0.07} & \textbf{0.47} & 0.15 & \textbf{799.17} & \textbf{11.36} & 256.98\\
\hline
\hline
\end{tabular}
\vspace{2mm}
\caption{EuRoC RMSE and NEES results for an estimator using IMU preintegration and approximating group Jacobians by $\eye$. Best RMSE (within 5\%) and NEES for each set are bold. Expected avg. NEES is shown in brackets.}
\label{tbl:euroc_results}
\vspace{-10mm}
\end{table*}





\bibliographystyle{IEEEtran}
\bibliography{IEEEabrv,paper}

\onecolumn
\newpage
\appendices

\markboth{Supplemental Material}
{Lisus \MakeLowercase{\textit{et al.}}: Consistency in Sliding Window Filtering with Unobservable States}

{\centering \LARGE \textbf{Supplementary Material} \par} \vspace{5mm}
The following supplementary material contains additional derivations related the null space computation procedure and Jacobian expressions. While the paper is self-contained without this material, the goal of this section is to make it easier for readers to apply the two requirements to their own algorithms, by explicitly showing how to analytically compute the null space from a known unobservable transformation. Appendix~\ref{sec:nullspace_explicit} explicitly writes out the null space computation procedure. It also presents one example of how to compute a null space state component, such as \eqref{eq:N_X_S} and \eqref{eq:N_X_R}, starting from the general form presented in \eqref{eq:N_definition}. 

Appendix~\ref{sec:Jacobians} presents expressions for the Jacobians necessary to conduct the consistency analysis of the considered sliding window filter in this letter. Specifically, the direct IMU integration error Jacobians, IMU preintegration error Jacobians, and visual error Jacobians with respect to the IMU states are all presented here. As the focus of the paper is implementing the two proposed consistency requirements, the prior error Jacobian and the visual error Jacobian with respect to the landmarks are not presented. The prior error is not used in the consistency analysis, whereas the visual error Jacobian with respect to the landmarks is multiplied through by the landmark component of the null space, which is always $\zero$. Additionally, bias components of state Jacobians are not explicitly defined, as their null space components are similarly always $\zero$. The goal of this section is to allow readers to verify for themselves the satisfaction of Requirement 1 for different representations, as presented in Table I. Full, detailed derivations can be found in \cite{Lisus2022}, albeit with a left-invariant orientation error definition for the separate group parametrization, as opposed to the right-invariant one used in this letter.

\section{VI-SLAM Unobservable Transformation and Null Space}\label{sec:nullspace_explicit}
Consider the full VI-SLAM state $\mathcal{X}^\mathrm{SLAM} = \left(\mathcal{X}_0, \ldots, \mathcal{X}_j, \mbf{z}^1, \ldots, \mbf{z}^L \right)$, where $\mathcal{X}_k$ is the IMU state at time $t = t_k$, parameterized either using the group $G^S$ or $G^R$, and $\mbf{z}^q \in \mathbb{R}^3$ is the anchored inverse depth parameterization of the $q^\mathrm{th}$ landmark. Consider the unobservable directions of the VI-SLAM problem given by $\mbs{\tau} = \begin{bmatrix} \phi_T & \mbf{r}_T^\trans \end{bmatrix}^\trans \in \mathbb{R}^4$. A transformation of the full SLAM state along these unobservable directions is defined as acting component-by-component, as 
    \begin{align}
        \mathcal{Y}^\mathrm{SLAM} = \mathcal{T}^\mathrm{SLAM}(\mathcal{X}^\mathrm{SLAM}, \mbs{\tau}) = \left(\mathcal{T}^X \left(\mathcal{X}_0, \mbs{\tau} \right), \ldots, \mathcal{T}^X \left(\mathcal{X}_k, \mbs{\tau} \right), \mathcal{T}^z \left(\mbf{z}^1, \mbs{\tau} \right), \ldots, \mathcal{T}^z \left(\mbf{z}^L, \mbs{\tau} \right) \right),
    \end{align}
where $\mathcal{Y}^\mathrm{SLAM}$ is the transformed full SLAM state, $\mathcal{T}^X \left(\cdot, \cdot \right)$ is the transformation function that transforms the IMU state, and $\mathcal{T}^z \left(\cdot, \cdot \right)$ is the transformation function that transforms the landmark parameters. Hence, the full null space of the VI-SLAM problem corresponding to the whole state is also computed term-by-term, as
\begin{align}
    \f{\partial\mbs{\eta} \left(\bar{\mathcal{X}}^\mathrm{SLAM}, \mathcal{T}^\mathrm{SLAM} \left(\bar{\mathcal{X}}^\mathrm{SLAM}, \mbs{\tau} \right) \right)}{\partial \mbs{\tau}}\biggr\rvert_{\mbsbar{\tau} = \zero} =
    \begin{bmatrix}
        \mbf{N}_{\mathcal{X}_0} \\
        \vdots \\
        \mbf{N}_{\mathcal{X}_j} \\
        \mbf{N}_{z^1} \\
        \vdots \\
        \mbf{N}_{z^L}
    \end{bmatrix},
\end{align}
where these components are individually defined as
    \begin{align}
        \mbf{N}_{\mathcal{X}_\imath} & = \f{\partial\mbs{\eta} \left(\bar{\mathcal{X}}_\imath, \mathcal{T}^X \left(\bar{\mathcal{X}}_\imath, \mbs{\tau} \right) \right)}{\partial \mbs{\tau}}\biggr\rvert_{\mbsbar{\tau} = \zero}, \hspace{5mm} \imath = 0, \ldots, j, \\
        \mbf{N}_{z^{\jmath}} & = \f{\partial\mbs{\eta} \left(\mbfbar{z}^\jmath, \mathcal{T}^z \left(\mbfbar{z}^\jmath, \mbs{\tau} \right) \right)}{\partial \mbs{\tau}}\biggr\rvert_{\mbsbar{\tau} = \zero}, \hspace{5mm} \jmath = 1, \ldots, L.
    \end{align}
As described in the letter, the landmark states are unaffected by the unobservable transformation. As such, the landmark state null space components are $\mbf{N}_{z^\jmath} = \mbf{0}, \hspace{3mm} \jmath = 1, \ldots, L$.

\subsection{Example Null Space Computation}
As an example of how to compute the null space component for the IMU state, consider the situation where the IMU state $\mathcal{X}$ is parameterized using the group $G^R$. First, consider the difference function between a state $\mathcal{X} = \scalebrack{ \mbf{X}^\mathrm{nav}, \bias } \in G^R$ and the transformed state 
\begin{align}
    \mathcal{Y}^X = \mathcal{T}^X (\mathcal{X}, \mbs{\tau}) = \scalebrack{ \mbf{Y}^\mathrm{nav}, \bias } \in G^R,
\end{align}
where $\mbs{\tau}=\bbm \phi_T & \pos_T^\trans \ebm^\trans \in \mathbb{R}^4$, and $\mbf{Y}^\mathrm{nav}$ is defined as
\begin{align}
    \mbf{Y}^{\mathrm{nav}} = \bbm \mbf{C}_T\mbf{C} & \mbf{C}_T\mbf{v} & \mbf{C}_T\mbf{r} + \mbf{r}_T\\ 0 & 1 & 0 \\ 0 & 0 & 1 \ebm = \bbm \mbf{C}_T & \zero & \mbf{r}_T\\ 0 & 1 & 0 \\ 0 & 0 & 1 \ebm \bbm \mbf{C} & \mbf{v} & \mbf{r}\\ 0 & 1 & 0 \\ 0 & 0 & 1 \ebm = \mbf{X}_T^\mathrm{nav}\mbf{X}^\mathrm{nav} \in SE_2(3).
\end{align}
The matrix $\mbf{X}_T^\mathrm{nav}$ can be written as 
    \begin{align}
        \mbf{X}_T^\mathrm{nav} = \Exp(\mbstilde{\tau}),
    \end{align}
where $\mbstilde{\tau} \left(\mbs{\tau} \right)$ is defined as
    \begin{align}
        \mbstilde{\tau} \left(\mbs{\tau} \right) = \bbm \mbf{g}\phi_T \\ \zero \\ \mbf{J}^{-1} \mbf{r}_T \ebm \in \mathbb{R}^9,
    \end{align}
where $\mbf{J}_\ell^{-1}$ is the inverse left Jacobian of $SO(3)$. The difference between the transformed and untransformed state is then
\begin{align}
    \mbs{\eta}^R(\mathcal{X}, \mathcal{Y}^X) = \begin{bmatrix} \Log(\mbf{X}^{\mathrm{nav}} {\mbf{Y}^\mathrm{nav}}^{-1}) \\ (\bias - \bias)\end{bmatrix} = \begin{bmatrix} \Log(\mbf{X}^{\mathrm{nav}} {\mbf{X}^\mathrm{nav}}^{-1}{\mbf{X}_T^\mathrm{nav}}^{-1}) \\ (\bias - \bias)\end{bmatrix} = \begin{bmatrix} -\mbstilde{\tau} \\ \zero\end{bmatrix}.
\end{align}
The null space, adopting the error definition $\mbs{\tau} = \mbsbar{\tau} - \delta\mbs{\tau}$, is then
\begin{align}
    \mbf{N}_{\mathcal{X}^R} = \f{\partial\mbs{\eta}^R\left(\mathcal{X}, \mathcal{Y}^X \right) }{\partial \mbs{\tau}}\biggr\rvert_{\mbs{\tau} = \zero} = \left. \f{\partial \begin{bmatrix} -\mbstilde{\tau}(\mbs{\tau}) \\ \zero\end{bmatrix} }{\partial \mbs{\tau}}\right\rvert_{\mbsbar{\tau} = \zero} = \bbm \grav & \zero\\ \zero & \zero \\ \zero & \eye \\ \zero & \zero \ebm.
\end{align}


\section{Error and Jacobian Expressions}\label{sec:Jacobians}
\subsection{Direct IMU Integration Error Jacobian}
The direct IMU error connects $\mathcal{X}_i$ to $\mathcal{X}_j$ using all measurements that occurred between $t_i$ and $t_j$, where $t_i$ and $t_j$ are consecutive timesteps at which camera measurements are recorded. Recall the recursive IMU process model given by 
\begin{align}\label{eq:recursive_f_imu_proc_model_supp}
   \mathcal{X}_j &= f_{j-1}\scalebrack{f_{j-2}\scalebrack{\mathcal{X}_{j-2}, \mbf{u}_{j-2}, \mbf{w}_{j-2}}, \mbf{u}_{j-1}, \mbf{w}_{j-1}} \\
&= f_{j-1}(\mathcal{X}_{j-1}, \mbf{u}_{j-1}, \mbf{w}_{j-1})\\
  &= f_{ij}(\mathcal{X}_i, u_{\imath:\jmath-1}, w_{\imath:\jmath-1}),
\end{align}
where $\mbf{u}_k \in \mathbb{R}^6$ is the IMU measurement at time $k$, $\mbf{w}_k \in \mathbb{R}^{12} $ is the stacked IMU measurement noise and bias random walk noise at time $k$, and ${x}_{i:j-1} = \{\mbf{x}_i, \cdots, \mbf{x}_{j-1}\}$ for $x \in [u, w]$. The direct IMU integration error is given by 
    \begin{align} 
        \mbf{e}_{u,j} = \mbs{\eta}(\check{\mathcal{X}}_j, \mathcal{X}_j),
    \end{align}
where $\check{\mathcal{X}}_j$ is computed by propagating $\mathcal{X}_i$ to time $t_j$ using \eqref{eq:recursive_f_imu_proc_model_supp}, and the appropriate definition for $\mbs{\eta} \left(\cdot, \cdot \right)$ is chosen based on the chosen IMU state parameterization.

To use this error in a sliding window filter, the Jacobian of the error with respect to both $\mathcal{X}_i$ and $\mathcal{X}_j$ is required. Defining the Jacobian of the process model~\eqref{eq:recursive_f_imu_proc_model_supp} with respect to $\mathcal{X}_i$ as
\begin{align}
   \f{D \check{\mathcal{X}}_j}{D \mathcal{X}_i} = \frac{D f_{ij}}{D {\mathcal{X}}_i},
\end{align}
the non-zero error Jacobian components are given by 
\begin{align}
    \f{D \mbf{e}_{u,j}}{D \mathcal{X}_i} &= \f{D \mbf{e}_{u,j}}{D \check{\mathcal{X}}_j}\f{D \check{\mathcal{X}}_j}{D \mathcal{X}_i} = -\mbs{\mathcal{J}}_\ell^{-1}(\mbfbar{e}_{u,j})\f{D \check{\mathcal{X}}_j}{D \mathcal{X}_i},\label{eq:e_u_jac_X_i_supp}\\
    \f{D \mbf{e}_{u,j}}{D \mathcal{X}_j} &= \mbs{\mathcal{J}}_\ell^{-1}(-\mbfbar{e}_{u,j})\label{eq:e_u_jac_X_j_supp},
\end{align}
with the left group Jacobian defined as ${\mbs{\mathcal{J}}_\ell^S}^{-1} = \diag(\mbf{J}_\ell^{-1}, \eye, \eye, \zero)$ and $\mbs{\mathcal{J}}_\ell^{R^{-1}} = \diag(\mbf{J}_\ell^{-1}, \zero)$ for $\mathcal{X}^S$ and $\mathcal{X}^R$ respectively, with $\mbf{J}_\ell$ either being the $SO(3)$ or $SE_2(3)$ group Jacobian respectively. 

The Jacobian $\frac{D \check{\mathcal{X}}_j}{D \mathcal{X}_i}$, representing the Jacobian of the process model with respect to $\mathcal{X}_i$, can be then computed for either $G^S$ or $G^R$ 
parametrization of the IMU state.

To determine the Jacobian of the process model $f_{ij}$ with respect to $\mathcal{X}_i$, it is first required to obtain the Jacobian of the one-step process model, $\mathcal{X}_k = f_{k-1} \left(\mathcal{X}_{k-1}, \mbf{u}_{k-1}, \mbf{w}_{k-1} \right)$, with respect to $\mathcal{X}_{k-1}$. This Jacobian can be found by linearizing the one-step process model as
    \begin{align}
        \delta \mbs{\xi}_k \approx \mbf{F}_{k-1} \delta \mbs{\xi}_{k-1},
    \end{align}
where the Jacobian $\mbf{F}_{k-1}$ is defined as
    \begin{align}
        \mbf{F}_{k-1} = \f{D \check{\mathcal{X}}_k}{D \mathcal{X}_\kmin} = \frac{D f_{k-1}}{D {\mathcal{X}}_\kmin}.
    \end{align}
Using these individual Jacobians from the one-step process model, the Jacobian of the recursive process model $f_{ij}$ can be computed as
\begin{align}
    \frac{D \check{\mathcal{X}}_j}{D \mathcal{X}_i} = \f{D {f}_{ij}}{D \mathcal{X}_i} &= \f{D \check{\mathcal{X}}_{j}}{D \check{\mathcal{X}}_{j-1}}\f{D \check{\mathcal{X}}_{j-1}}{D \check{\mathcal{X}}_{j-2}}\hdots\f{D \check{\mathcal{X}}_{i+2}}{D \check{\mathcal{X}}_{i+1}}\f{D \check{\mathcal{X}}_{i+1}}{D \mathcal{X}_i}\\
    &= \f{D f_{j-1}}{D f_{j-2}}\f{D f_{j-2}}{D f_{j-3}}\hdots\f{D f_{i+1}}{D f_{i}}\f{D f_{i}}{D \mathcal{X}_i}.\label{eq:full_jac_mult}
\end{align}

If $\mathcal{X}$ is parameterized using the separate group representation, such that $\mathcal{X} \in G^S$, the Jacobian of the one-step process model $\mbf{F}_{k-1}^{S}$ is written as

\begin{align}
    \mbf{F}_\kmin^S = 
        \begin{bmatrix} 
                \eye & \zero & \zero & \star\\
                - T_\kmin(\dcmbar_\kmin(\mbf{u}_\kmin^a - \babar_\kmin))^\wedge & \eye & \zero & \star\\
                - \onehalf T_\kmin^2(\dcmbar_\kmin(\mbf{u}_\kmin^a - \babar_\kmin))^\wedge & T_\kmin\eye & \eye & \star \\
                \zero & \zero & \zero & \eye\\
        \end{bmatrix},
\end{align}
where $T_\kmin = t_k - t_{k-1}$ and $\star$ denote the bias-related $3\times6$ non-zero blocks.

Alternatively, if $\mathcal{X}$ is parameterized by grouping the navigation state into an element of $SE_2(3)$, such that $\mathcal{X} \in G^R$, the Jacobian of the one-step process model is written as
\begin{align}
    \mbf{F}_\kmin^R &= \bbm 
                \eye & \zero & \zero & \star\\
                T_\kmin\grav^\wedge & \eye & \zero & \star \\
                \onehalf T_\kmin^2\grav^\wedge & T_\kmin\eye & \eye & \star\\
                \zero & \zero & \zero & \eye\ebm,
\end{align}
where again $T_\kmin = t_k - t_{k-1}$, $\star$ denote the bias-related $3\times6$ non-zero blocks.

The final $\frac{D \check{\mathcal{X}}_j}{D \mathcal{X}_i}$ Jacobian in \eqref{eq:e_u_jac_X_i_supp} is computed by multiplying the one-step process model Jacobians as in \eqref{eq:full_jac_mult}. Requirement 1 is satisfied for this error if
\begin{align}
    \f{D \mbf{e}_{u,j}}{D \mathcal{X}_i}\mbf{N}_{\mathcal{X}_i} + \f{D\mbf{e}_{u,j}}{D \mathcal{X}_j}\mbf{N}_{\mathcal{X}_j} = \zero \label{eq:imu_dir_HN}
\end{align}
is satisfied for a given parameterization of the IMU state. Multiplying out the one-step Jacobians per \eqref{eq:full_jac_mult} can be tedious, however it can be verified that \eqref{eq:imu_dir_HN} will always analytically cancel to $\zero$ if the group Jacobians in~\eqref{eq:e_u_jac_X_i_supp} and~\eqref{eq:e_u_jac_X_j_supp} are approximated by $\eye$. Note, the multiplied out result for the $\mathcal{X}^R$ parametrization presented in the paper in~\eqref{eq:explicit_Jacobian} makes use of the sum of squares identity
\begin{align}
	\sum_{k=i}^{j-1}T_k^2 = \left(\sum_{k=i}^{j-1}T_k\right)^2 - 2\sum_{k=i}^{j-1}\left(T_k\sum_{k_2=i}^{k-1}T_k\right) = (\Delta t_{ij})^2 - 2\sum_{k=i}^{j-1}(T_k\Delta t_{ik}),
\end{align}
with $\Delta t_{ij} = \sum_{k=i}^{j-1} T_k$.

\subsection{IMU Preintegration Error Jacobians}
Consider the RMI $\RMIX_X$, which connects $\mathcal{X}_i$ to $\mathcal{X}_j$, and which is either an element of $G^S$ or $G^R$. The individual components are computed from the expressions
\begin{align}
	\Delta\mbf{C}_{X_{ij}} &\triangleq \dcmt_i\dcm_j \in SO(3) ,\label{eq:ct_rmi_def_dcm_X_supp} \\
	\Delta\mbf{v}_{X_{ij}} &\triangleq \dcmt_i(\vel_j - \vel_i - \mbf{g}\Delta t_{ij}) \in \mathbb{R}^3, \label{eq:ct_rmi_def_vel_X_supp}\\
	\Delta\mbf{r}_{X_{ij}} &\triangleq \dcmt_i(\pos_j - \pos_i - \vel_i\Delta t_{ij} - \frac{1}{2}\mbf{g}\Delta t_{ij}^2) \in \mathbb{R}^3 ,\label{eq:ct_rmi_def_pos_X_supp}\\
	\Delta\mbf{b}_{X_{ij}} &\triangleq \bias_j - \bias_i \in \mathbb{R}^3 \label{eq:ct_rmi_def_b_X_supp},
\end{align}
where $\Delta t_{ij} = \sum_{k=i}^{j-1} T$.
Next, consider the RMI $\RMIX_Y$, again either an element of $G^S$ or $G^R$, where the individual components are computed incrementally from new arriving IMU measurements as
\begin{align}
	\Delta\mbf{C}_{Y_{ij}} &\triangleq \prod_{k=i}^{j-1}\Exp\left((\mbf{u}_k^g - \bg_k - \mbf{w}_k^g) T\right) \in SO(3),\label{eq:dt_rmi_def_dcm_Y_supp}\\
	\Delta\mbf{v}_{Y_{ij}} &\triangleq \sum_{k=i}^{j-1}\Delta\mbf{C}_{Y_{ik}}(\mbf{u}_k^a - \ba_k - \mbf{w}_k^a) T \in \mathbb{R}^3,\label{eq:dt_rmi_def_v_Y_supp}\\
	\Delta\mbf{r}_{Y_{ij}} &\triangleq \sum_{k=i}^{j-1}\Delta\mbf{v}_{ik}T + \frac{1}{2}\Delta\mbf{C}_{Y_{ik}}(\mbf{u}_k^a - \ba_k - \mbf{w}_k^a) T^2 \in \mathbb{R}^3 \label{eq:dt_rmi_def_r_Y_supp},\\
	\Delta\mbf{b}_{Y_{ij}} &\triangleq \mbf{0} \in \mathbb{R}^3 \label{eq:dt_rmi_def_b_Y_supp}.
\end{align}
An error between $\RMIX_X$ and $\RMIX_Y$ can be computed as 
    \begin{align}
        \mbf{e}_{\RMIX_{ij}} = \mbs{\eta} \left(\RMIX_Y, \RMIX_X \right),
    \end{align}
where the form of the function $\mbs{\eta} \left(\cdot, \cdot \right)$ depends on the chosen IMU state parameterization. Similarly to using direct IMU integration, the Jacobians of the error $\mbf{e}_{\RMIX_{ij}}$ with respect to both $\mathcal{X}_i$ and $\mathcal{X}_j$ are required. The form of all Jacobians depends on the chosen IMU state parameterization and RMI parameterization. The RMI parameterization is always chosen to be consistent with the IMU state parameterization. Specifically, when the IMU state is parameterized as $\mathcal{X}^S \in G^S$, the RMIs are parametrized as $\RMIX_X^S \in G^S$ and $\RMIX_Y^S \in G^S$. Conversely, when the IMU state is parameterized as $\mathcal{X}^R \in G^R$, the RMIs are parameterizd as $\RMIX_X^R \in G^R$ and $\RMIX_X^R \in G^R$.

These desired Jacobians are written using the chain rule as
\begin{align}
    \f{D \mbf{e}_{\RMIX_{ij}}}{D \mathcal{X}_i} &= \f{D \mbf{e}_{\RMIX_{ij}}}{D \RMIX_Y}\f{D \RMIX_Y}{D \mathcal{X}_i} + \f{D \mbf{e}_{\RMIX_{ij}}}{D \RMIX_X}\f{D \RMIX_X}{D \mathcal{X}_i} \nonumber\\ &= -\mbs{\mathcal{J}}_\ell^{-1}(\mbf{e}_{\RMIX_{ij}})\f{D \RMIX_Y}{D \mathcal{X}_i}  + \mbs{\mathcal{J}}_\ell^{-1}(-\mbf{e}_{\RMIX_{ij}})\f{D \RMIX_X}{D \mathcal{X}_i},\label{eq:ahh_1}\\
    \f{D \mbf{e}_{\RMIX_{ij}}}{D \mathcal{X}_j} &= \f{D \mbf{e}_{\RMIX_{ij}}}{D \RMIX_Y}\f{D \RMIX_Y}{D \mathcal{X}_j} + \f{D \mbf{e}_{u,j}}{D \RMIX_X}\f{D \RMIX_X}{D \mathcal{X}_j} \nonumber\\ &= -\mbs{\mathcal{J}}_\ell^{-1}(\mbf{e}_{\RMIX_{ij}})\f{D \RMIX_Y}{D \mathcal{X}_j} + \mbs{\mathcal{J}}_\ell^{-1}(-\mbf{e}_{\RMIX_{ij}})\f{D \RMIX_X}{D \mathcal{X}_j}\label{eq:ahh_2},
\end{align}
with the left group Jacobian defined as ${\mbs{\mathcal{J}}_\ell^S}^{-1} = \diag(\mbf{J}_\ell^{-1}, \eye, \eye, \zero)$ and $\mbs{\mathcal{J}}_\ell^{R^{-1}} = \diag(\mbf{J}_\ell^{-1}, \zero)$ for $\mathcal{X}^S$ and $\mathcal{X}^R$ respectively, with $\mbf{J}_\ell$ either being the $SO(3)$ or $SE_2(3)$ group Jacobian. The remainder of the Jacobians in \eqref{eq:ahh_1} and \eqref{eq:ahh_2} are presented below.

The linearization of $\RMIX_Y$ with respect to the states $\mathcal{X}_i$ and $\mathcal{X}_j$ yields two Jacobians that have the same form, albeit with different $\star$ values, for both IMU state parametrizations:
\begin{align}
    \f{D \RMIX_Y}{D \mathcal{X}_i} &= 
    \bbm 
        \zero & \zero & \zero & \star \\
        \zero & \zero & \zero & \star \\
        \zero & \zero & \zero & \star \\ 
        \zero & \zero & \zero & \zero\ebm, \\
    \f{D \RMIX_Y}{D \mathcal{X}_j} & = \mbf{0}.
\end{align}

The linearization of $\RMIX_X$ with respect to the states $\mathcal{X}_i$ and $\mathcal{X}_j$ is more parametrization-specific. For the separate group parametrization, the Jacobians are
\begin{align}
    \f{D \RMIX_X^R}{D \mathcal{X}_i^R} &= \bbm -\dcmbart_i & \zero & \zero & \zero \\
    \dcmbart_i(\dcmbar_i\RMIvbar_X)^\wedge & -\dcmbart_i & \zero & \zero \\
    \dcmbart_i(\dcmbar_i\RMIrbar_X)^\wedge & -\Deltat_{ij}\dcmbart_i & -\dcmbart_i & \zero \\ 
    \zero & \zero & \zero & -\eye \ebm,\\
    \f{D \RMIX_X^R}{D \mathcal{X}_j^R} &= \bbm \dcmbart_i & \zero & \zero & \zero \\
    \zero & \dcmbart_i & \zero & \zero \\
    \zero & \zero & \dcmbart_i & \zero \\ 
    \zero & \zero & \zero & \eye\ebm,
\end{align}
where $\Delta t_{ij} = \sum_{k=i}^{j-1} T_k$.

For the $SE_2(3)$ group parametrization, the Jacobians are
\begin{align}
    \f{D \RMIX_X^R}{D \mathcal{X}_i^R} &= \bbm -\dcmbart_i & \zero & \zero & \zero \\
    \dcmbart_i\vbar_i^\wedge & -\dcmbart_i & \zero & \zero \\
    \dcmbart_i(\rbar_i + \vbar_i\Deltat_{ij})^\wedge & -\Deltat_{ij}\dcmbart_i & -\dcmbart_i & \zero \\ 
    \zero & \zero & \zero & -\eye \ebm,\\
    \f{D \RMIX_X^R}{D \mathcal{X}_j^R} &= \bbm \dcmbart_i & \zero & \zero & \zero \\
    -\dcmbart_i\scalebrack{\vbar_i + \grav\Deltat_{ij}}^\wedge & \dcmbart_i & \zero & \zero \\
    -\dcmbart_i\scalebrack{\rbar_i + \vbar_i\Deltat_{ij} + \onehalf\grav\Deltat_{ij}^2}^\wedge & \zero & \dcmbart_i & \zero \\ 
    \zero & \zero & \zero & \eye\ebm,
\end{align}
where $\Delta t_{ij} = \sum_{k=i}^{j-1} T_k$.

Requirement 1 is satisfied for this error if
\begin{align}
    \f{D \mbf{e}_{\RMIX_{ij}}}{D \mathcal{X}_i}\mbf{N}_{\mathcal{X}_i} + \f{D \mbf{e}_{\RMIX_{ij}}}{D \mathcal{X}_j}\mbf{N}_{\mathcal{X}_j} = \zero \label{eq:RMI_HN}
\end{align}
is satisfied for a given parameterization of the IMU state. This multiplication is much easier to do than the one involved with the direct IMU integration, and can be seen to be upheld for both considered parameterizations of the IMU state. Additionally, unlike with using direct IMU integration, Requirement 1 is satisfied for this error with, or without, the approximation of group Jacobians in~\eqref{eq:ahh_1} and~\eqref{eq:ahh_2}.

\subsection{Visual Error Jacobians}
Visual errors are added to the sliding window filter for each measurement of each visible landmark in the window.
Consider the measurement of a landmark $\ell^q$, taken from camera $\imath$ at at time $t = t_k$, written as $\mbf{y}_{q, \imath, k}$. The visual error $\mbf{e}_{q, \imath, k}$ compares this measurement to the measurement model, written as 
    \begin{align} \label{eq:visual_error}
        \mbf{e}_{q, \imath, k} = \mbf{y}_{q, \imath, k} - \mbf{g}(\mbf{r}_{c_k^\imath}^{q c_k^{\imath}}).
    \end{align}
The function $\mbf{g} (\cdot)$ is the pinhole camera model given by
\begin{align}\label{eq:norm_pix_coord_proj_function_supp}
	\mbf{g}(\mbf{r}_{c_k^\imath}^{q c_k^{\imath}}) = \begin{bmatrix}
	x_n^q\\ y_n^q\end{bmatrix} = \begin{bmatrix} x^q / z^q & y^q / z^q
	\end{bmatrix}^\trans,
\end{align}
where $\mbf{r}_{c_k^\imath}^{q c_k^\imath}$ is written, omitting the camera index $\imath$ and defining $\dcm_{ab_k} = \dcm_k$, as
\begin{align}\label{eq:p_a_to_p_c_projection_supp}
    \mbf{r}_{c}^{q c} = \begin{bmatrix} x^q \\ y^q \\ z^q \end{bmatrix} = \dcmt_{b c} \scalebrack{\dcmt_k (\mbf{r}_a^{qw} - \pos_a^{z_kw}) - \mbf{t}_b^{c z}}.
\end{align}
Finally, to write the measurement model as a function of the states to be estimated, the inertial landmark position $\mbf{r}_a^{qw}$ need to be written as a function of the anchored inverse depth parameterization, $\mbf{z}^q = \bbm \alpha_c^q & \beta_c^q & \lambda_c^q \ebm^\trans \in \mathbb{R}^3$, as
    \begin{align}\label{eq:world_frame_rep_from_3invdep_supp}
    	\mbf{r}_a^{qw} = \mbf{C}_i\scalebrack{\mbf{C}_{bc}\frac{1}{\lambda_c^q}\begin{bmatrix}\alpha_c^q \\ \beta_c^q \\ 1 \end{bmatrix} + \mbf{t}_{b}^{c z}} + \mbf{r}_i,
    \end{align}
where $\mbf{C}_i \in SO(3)$ and $\mbf{r}_i \in \mathbb{R}^3$ are the anchoring IMU orientation and position. 
The error term in \eqref{eq:visual_error} is a function of the landmark parameters $\mbf{z}^{q}$, the anchoring IMU state $\mathcal{X}_i$, and the IMU state at the measurement time $\mathcal{X}_k$. As with previous error terms, the IMU states can either be parameterized using the group $G^S$ or $G^R$. To use this error term within a sliding window filter, the individual Jacobians of the error with respect to $\mbf{z}_{c}^q$, $\mathcal{X}_i$, and $\mathcal{X}_k$, are required. However, since the null space component associated with the landmark will always be $\zero$, this Jacobian is not provided here. The other individual Jacobians are written using the chain rule as
    \begin{align} \label{eq:jac_inv_depth}
        \f{D \mbf{e}_{y,\imath, k}}{D \mathcal{X}_k} &= -\f{\partial \mbf{g}}{\partial \mbf{r}_c^{qc}} \f{D \mbf{r}_c^{qc}}{D \mathcal{X}_k},\\ \label{eq:visual_jac_x_i}
        \f{D \mbf{e}_{y, \imath, k}}{D \mathcal{X}_i} &= -\f{\partial \mbf{g}}{\partial \mbf{r}_c^{qc}} \f{\partial \mbf{r}_c^{qc}}{\partial \mbf{r}_a^{qw}} \f{D \mbf{r}_a^{qw}}{D \mathcal{X}_i}.
    \end{align}
The projection function Jacobian, $\f{\partial \mbf{g}}{\partial \mbf{r}_c^{qc}}$ is needed in each visual error Jacobian, and is given by 
\begin{align}\label{eq:jac_projection_function}
    \f{\partial \mbf{g}}{\partial \mbf{r}_c^{qc}} = \f{1}{z^q}
    \bbm 1 & 0 & -\f{x^q}{z^q} \\
    0 & 1 & -\f{y^q}{z^q}
    \ebm,
\end{align}
where $\mbf{r}_c^{qc} = \begin{bmatrix} x^q & y^q & z^q \end{bmatrix}^\trans$. The Jacobian $\f{\partial \mbf{r}_c^{qc}}{\partial \mbf{r}_a^{qw}}$ is given by 
    \begin{align}\label{eq:visual_jacobians}
        \f{\partial \mbf{r}_c^{qc}}{\partial \mbf{r}_a^{qw}} = \dcmt_{bc} \bar{\dcm}_{k}^\trans.
    \end{align}
    
The next Jacobian required is $\f{D \mbf{r}_c^{qc}}{D \mathcal{X}_k}$, which depends on the chosen parameterization for $\mathcal{X}_k$. Denoting $\mathcal{X}_k^S \in G^S$ and $\mathcal{X}_k^R \in G^R$, the Jacobians for both these cases are given by 
\begin{align}
    \f{D \mbf{r}_c^{qc}}{D \mathcal{X}_k^S} &= \bbm - \dcmt_{bc}\mbfbar{C}_{k}^\trans(\mbfbar{r}_a^{qw} - \mbf{r}_a^{z_k w})^\wedge & \zero & \dcmt_{bc}\dcmbar_k^\trans & \zero  \ebm,\label{eq:euc_p_partial_X_S}\\
    \f{D \mbf{r}_c^{qc}}{D \mathcal{X}_k^R} &= \bbm - \dcmt_{bc}\mbfbar{C}_{k}^\trans\mbfbar{r}_a^{qw^\wedge} & \zero & \dcmt_{bc}\mbfbar{C}_{k}^\trans & \zero \ebm,\label{eq:euc_p_partial_X_R}
\end{align}
Finally, the last Jacobian in the chain rule is $\f{\partial \mbf{r}_a^{qw}}{\partial \mathcal{X}_i}$, and the form of this Jacobian once again depends on the chosen parameterization for $\mathcal{X}_i$. Again denoting $\mathcal{X}_i^S \in G^S$ and $\mathcal{X}_i^R \in G^R$, the Jacobians for both these cases are given by 
\begin{align}\label{eq:partial_f_partial_Xi_S_inv1}
	\frac{D \mbf{r}_a^{qw}}{D \mathcal{X}_i^S} & = 
    \begin{bmatrix}
	\left( \mbfbar{C}_{i}\scalebrack{\mbf{C}_{bc}\frac{1}{{\lambda}}\begin{bmatrix} \bar{\alpha} \\ \bar{\beta} \\ 1 \end{bmatrix} + \mbf{t}_{b}^{cz}}\right)^\wedge & \mbf{0} & -\eye & \mbf{0}
	\end{bmatrix}, \\
 	\frac{D \mbf{r}_a^{qw}}{D \mathcal{X}_i^R} & =
    \begin{bmatrix}
	\scalebrack{\mbfbar{C}_{i} \scalebrack{\mbf{C}_{bc}\frac{1}{{\lambda}}\begin{bmatrix} \bar{\alpha} \\ \bar{\beta} \\ 1 \end{bmatrix} + \mbf{t}_{b}^{cz}} + \mbfbar{r}_a^{z_iw}}^\wedge & \mbf{0} & -\eye & \mbf{0}
	\end{bmatrix}.
\end{align}

Requirement 1 is satisfied for visual errors if 
    \begin{align}
        \f{D \mbf{e}_{y,\imath, k}}{D \mathcal{X}_k} \mbf{N}_{\mathcal{X}_k} + \f{D \mbf{e}_{y, \imath, k}}{D \mathcal{X}_i} \mbf{N}_{\mathcal{X}_i} + \f{D \mbf{e}_{y, \imath, k}}{D \mbf{z}^{q}} \mbf{N}_{z^q} = \mbf{0}
    \end{align}
is satisfied for a given parameterization of the IMU state. Note that the unobservable transformation of the state leaves the landmark parameters unchanged, such that $\mathcal{T}^z \left(\mbf{z}^q, \mbs{\tau} \right) = \mbf{z}^q$, and hence, the null space component corresponding to the landmark parameters is $\mbf{N}_{z^q} = \mbf{0}$. Hence, the null space equation simplifies to 
\begin{align}
    \f{D \mbf{e}_{y,\imath, k}}{D \mathcal{X}_k} \mbf{N}_{\mathcal{X}_k} + \f{D \mbf{e}_{y, \imath, k}}{D \mathcal{X}_i} \mbf{N}_{\mathcal{X}_i} = \mbf{0},
\end{align}
which can be shown to be satisfied for both considered parameterizations of the IMU state.

\end{document}